\documentclass{article}

\usepackage[preprint]{neurips_2024}

\usepackage[utf8]{inputenc} 
\usepackage[T1]{fontenc}    
\usepackage{hyperref}       
\usepackage{url}            
\usepackage{booktabs}       
\usepackage{amsfonts}       
\usepackage{nicefrac}       
\usepackage{microtype}      
\usepackage{xcolor}         
\usepackage{svg}
\usepackage{amsmath}
\usepackage{multirow}
\usepackage{graphicx}
\usepackage{subcaption}

\title{Transfusion: Predict the Next Token and\\Diffuse Images with One Multi-Modal Model}

\author{
\textbf{Chunting Zhou}$^{\mu}$\thanks{Equal contribution.} \qquad
\textbf{Lili Yu}$^{\mu*}$ \qquad
\textbf{Arun Babu}$^{\delta}$\thanks{Work done while at Meta.} \qquad
\textbf{Kushal Tirumala}$^{\mu}$ \\
\textbf{Michihiro Yasunaga}$^{\mu}$ \qquad
\textbf{Leonid Shamis}$^{\mu}$ \qquad
\textbf{Jacob Kahn}$^{\mu}$ \qquad
\textbf{Xuezhe Ma}$^{\sigma}$ \\
\textbf{Luke Zettlemoyer}$^{\mu}$ \qquad
\textbf{Omer Levy}$^{\dagger}$ \\
\\
$^\mu$ Meta \\
$^\delta$ Waymo
$^\sigma$ University of Southern California
}

\begin{document}

\maketitle

\begin{abstract}
We introduce Transfusion, a recipe for training a multi-modal model over discrete and continuous data.
Transfusion combines the language modeling loss function (next token prediction) with diffusion to train a single transformer over mixed-modality sequences.
We pretrain multiple Transfusion models up to 7B parameters from scratch on a mixture of text and image data, establishing scaling laws with respect to a variety of uni- and cross-modal benchmarks.
Our experiments show that Transfusion scales significantly better than quantizing images and training a language model over discrete image tokens.
By introducing modality-specific encoding and decoding layers, we can further improve the performance of Transfusion models, and even compress each image to just 16 patches.
We further demonstrate that scaling our Transfusion recipe to 7B parameters and 2T multi-modal tokens produces a model that can generate images and text on a par with similar scale diffusion models and language models, reaping the benefits of both worlds.
\end{abstract}
\section{Introduction}

Multi-modal generative models need to be able to perceive, process, and produce both discrete elements (such as text or code) and continuous elements (e.g. image, audio, and video data).
While language models trained on the next token prediction objective dominate discrete modalities \citep{gpt4, llama3}, diffusion models \citep{ho2020denoising, ldm} and their generalizations \citep{lipman2022flow} are the state of the art for generating continuous modalities \citep{dai2023emu, sd3, bar2024lumiere}.
Many efforts have been made to combine these approaches, including extending a language model to use a diffusion model as a tool, either explicitly \citep{liu2023llavaplus} or by grafting a pretrained diffusion model onto the language model \citep{dong2023dreamllm,koh2024generating}.
Alternatively, one can quantize the continuous modalities \citep{vqvae} and train a standard language model over discrete tokens \citep{dalle, yu2022scaling, yu2023scaling}, simplifying the model's architecture at the cost of losing information.
In this work, we show it is possible to fully integrate both modalities, with no information loss, by training a single model to both predict discrete text tokens and diffuse continuous images.

\begin{figure}
\centering
\includegraphics[width=\linewidth]{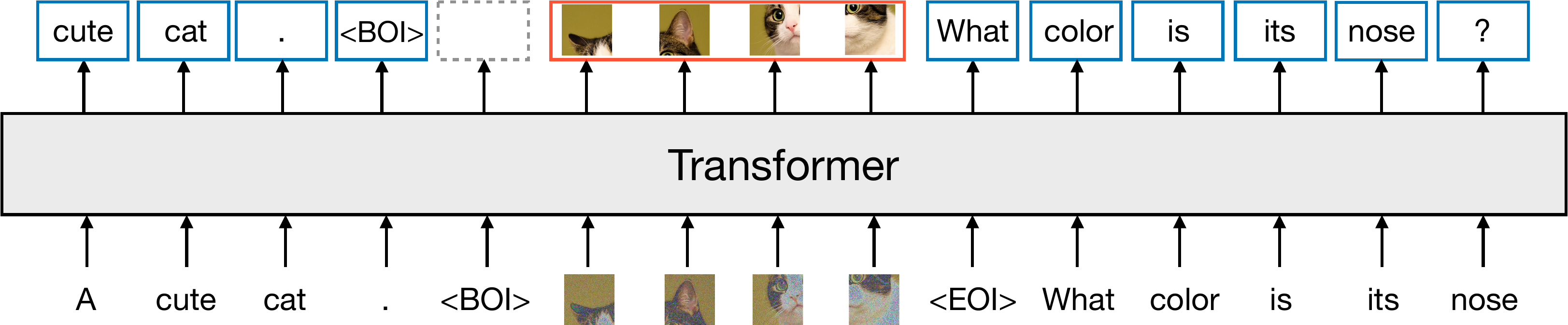}
\caption{A high-level illustration of Transfusion. A single transformer perceives, processes, and produces data of every modality. Discrete (text) tokens are processed autoregressively and trained on the \textcolor[HTML]{2171c7}{next token prediction} objective. Continuous (image) vectors are processed together in parallel and trained on the \textcolor[HTML]{e69500}{diffusion} objective. Marker BOI and EOI tokens separate the modalities.}
\label{fig:transfusion}
\end{figure}

We introduce \textbf{Transfusion}, a recipe for training a model that can seamlessly generate discrete and continuous modalities.
We demonstrate Transfusion by pretraining a transformer model on 50\% text and 50\% image data using a different objective for each modality: next token prediction for text and diffusion for images.
The model is exposed to both modalities and loss functions at each training step.
Standard embedding layers convert text tokens to vectors, while patchification layers represent each image as a sequence of patch vectors.
We apply causal attention for text tokens and bidirectional attention for image patches.
For inference, we introduce a decoding algorithm that combines the standard practices of text generation from language models and image generation from diffusion models.
Figure~\ref{fig:transfusion} illustrates Transfusion.

In a controlled comparison with Chameleon's discretization approach \citep{team2024chameleon}, we show that Transfusion models scale better in every combination of modalities.
In text-to-image generation, we find that Transfusion exceeds the Chameleon approach at less than a third of the compute, as measured by both FID and CLIP scores.
When controlling for FLOPs, Transfusion achieves approximately 2$\times$ lower FID scores than Chameleon models.
We observe a similar trend in image-to-text generation, where Transfusion matches Chameleon at 21.8\% of the FLOPs.
Surprisingly, Transfusion is also more efficient at learning text-to-text prediction, achieving perplexity parity on text tasks around 50\% to 60\% of Chameleon's FLOPs.

Ablation experiments reveal critical components and potential improvements for Transfusion.
We observe that the intra-image bidirectional attention is important, and that replacing it with causal attention hurts text-to-image generation.
We also find that adding U-Net down and up blocks to encode and decode images enables Transfusion to compress larger image patches with relatively small loss to performance, potentially decreasing the serving costs by up to 64$\times$.

\begin{figure}[htp]
    \centering
    \subfloat[An armchair in the shape of an avocado]{\includegraphics[width=0.23\textwidth]{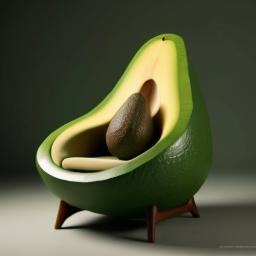}}\hfill
    \subfloat[A bread, an apple, and a knife on a table]{\includegraphics[width=0.23\textwidth]{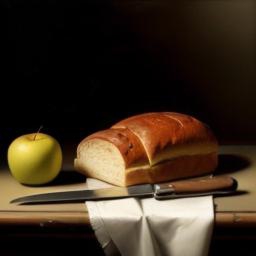}}\hfill
    \subfloat[A corgi.]{\includegraphics[width=0.23\textwidth]{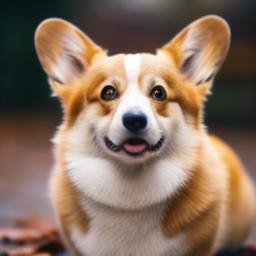}}\hfill
    \subfloat[human life depicted entirely out of fractals]{\includegraphics[width=0.23\textwidth]{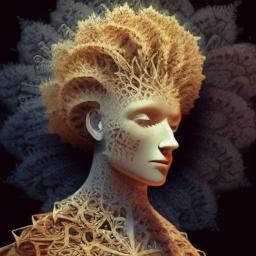}}\\[10pt]
    
    \subfloat[A blue jay standing on a large basket of rainbow macarons.]{\includegraphics[width=0.23\textwidth]{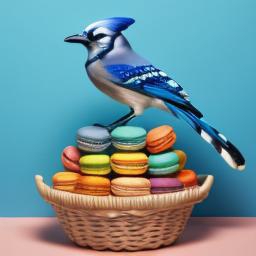}}\hfill
    \subfloat[``Transfusion" is written on the blackboard.]{\includegraphics[width=0.23\textwidth]{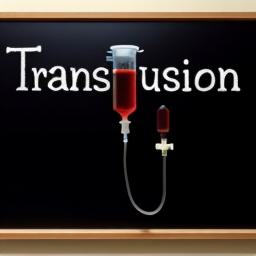}}\hfill
    \subfloat[A close up photo of a human hand, hand model. High quality]{\includegraphics[width=0.23\textwidth]{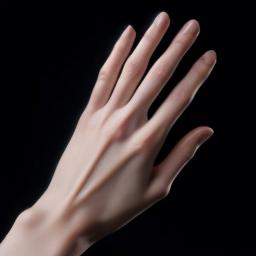}}\hfill
    \subfloat[A cloud in the shape of two bunnies playing with a ball. The ball is made of clouds too.]{\includegraphics[width=0.23\textwidth]{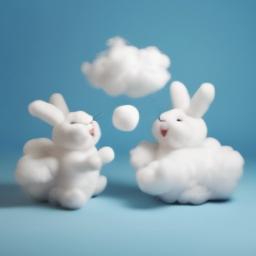}}\\[10pt]
    
    \subfloat[the word `START' on a blue t-shirt]{\includegraphics[width=0.23\textwidth]{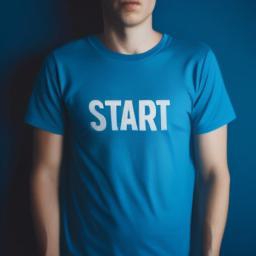}}\hfill
    \subfloat[A Dutch still life of an arrangement of tulips in a fluted vase. The lighting is subtle, casting gentle highlights on the flowers and emphasizing their delicate details and natural beauty.]{\includegraphics[width=0.23\textwidth]{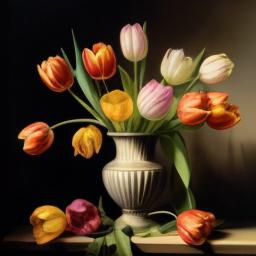}}\hfill
    \subfloat[A wall in a royal castle. There are two paintings on the wall. The one on the left a detailed oil painting of the royal raccoon king. The one on the right a detailed oil painting of the royal raccoon queen.]{\includegraphics[width=0.23\textwidth]{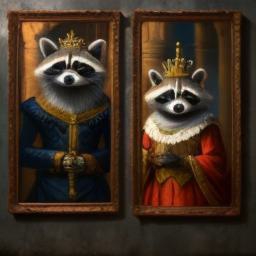}}\hfill
    \subfloat[Three spheres made of glass falling into ocean. Water is splashing. Sun is setting.]{\includegraphics[width=0.23\textwidth]{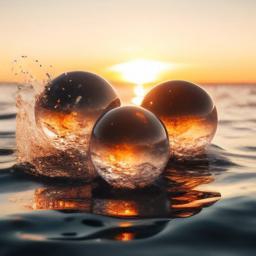}}\\[10pt]
    
    \subfloat[A transparent sculpture of a duck made out of glass.]{\includegraphics[width=0.23\textwidth]{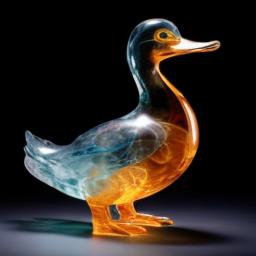}}\hfill
    \subfloat[A chromeplated cat sculpture placed on a Persian rug.]{\includegraphics[width=0.23\textwidth]{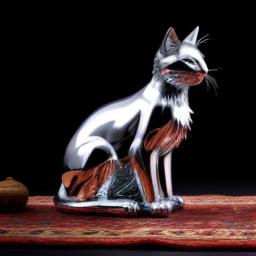}}\hfill
    \subfloat[A kangaroo holding a beer, wearing ski goggles and passionately singing silly songs.]{\includegraphics[width=0.23\textwidth]{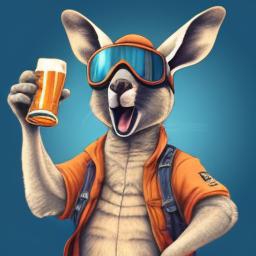}}\hfill
    \subfloat[an egg and a bird made of wheat bread]{\includegraphics[width=0.23\textwidth]{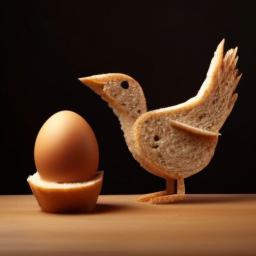}}\\[10pt]
\caption{Generated images from a 7B Transfusion trained on 2T multi-modal tokens.}
\label{fig:samples1}
\end{figure}

Finally, we demonstrate that Transfusion can generate images at similar quality to other diffusion models.
We train from scratch a 7B transformer enhanced with U-Net down/up layers (0.27B parameters) over 2T tokens: 1T text tokens, and approximately 5 epochs of 692M images and their captions, amounting to another 1T patches/tokens.
Figure~\ref{fig:samples1} shows some generated images sampled from the model.
On the GenEval \citep{ghosh2023geneval} benchmark, our model outperforms other popular models such as DALL-E 2 and SDXL; unlike those image generation models, it can generate text, reaching the same level of performance as Llama 1 on text benchmarks.
Our experiments thus show that Transfusion is a promising approach for training truly multi-modal models.

\captionsetup[subfigure]{labelformat=empty}

\section{Background}
\label{sec:background}

Transfusion is a single model trained with two objectives: language modeling and diffusion.
Each of these objectives represents the state of the art in discrete and continuous data modeling, respectively. 
This section briefly defines these objectives, as well as background on latent image representations.

\subsection{Language Modeling}
\label{sec:lm}

Given a sequence of discrete tokens $y = y_1, ..., y_n$ from a closed vocabulary $V$, a language model predicts the probability of the sequence $P(y)$.
Standard language models decompose $P(y)$ into a product of conditional probabilities $\prod_{i=1}^n P_\theta(y_i|y_{<i})$.
This creates an autoregressive classification task, where the probability distribution of each token $y_i$ is predicted conditioned on the prefix of a sequence $y_{<i}$ using a single distribution $P_\theta$ parameterized by $\theta$.
The model can be optimized by minimizing the cross-entropy between $P_\theta$ and the empirical distribution of the data, yielding the standard next-token prediction objective, colloquially referred to as \textit{LM loss}:
\begin{equation}
\mathcal{L}_{\text{LM}} = \mathbb{E}_{y_i} \big[ - \log P_\theta (y_i | y_{<i}) \big]
\end{equation}
Once trained, language models can also be used to generate text by sampling token by token from the model distribution $P_\theta$, typically using temperature and top-p truncation.

\subsection{Diffusion}
\label{sec:diffusion}

Denoising diffusion probabilistic models (a.k.a. \textit{DDPM} or \textit{diffusion models}) operate on the principle of learning to reverse a gradual noise-addition process \citep{ho2020denoising}.
Unlike language models that typically work with discrete tokens ($y$), diffusion models operate over continuous vectors ($\mathbf{x}$), making them particularly suited for tasks involving continuous data like images.
The diffusion framework involves two processes: a forward process that describes how the original data is turned into noise, and a reverse process of denoising that the model learns to perform.

\paragraph{Forward Process}
From a mathematical perspective, the forward process defines how the noised data (which serves as the model input) is created.
Given a data point $\mathbf{x}_0$, \citet{ho2020denoising} define a Markov chain that gradually adds Gaussian noise over $T$ steps, creating a sequence of increasingly noisy versions $\mathbf{x}_1, \mathbf{x}_2, ..., \mathbf{x}_T$.
Each step of this process is defined by $q(\mathbf{x}_t | \mathbf{x}_{t-1}) = \mathcal{N}(\mathbf{x}_t; \sqrt{1 - \beta_t}\mathbf{x}_{t-1}, \beta_t\mathbf{I})$, where $\beta_t$ increases over time according to a predefined noise schedule (see below).
This process can be reparameterized in a way that allows us to directly sample $\mathbf{x}_t$ from $\mathbf{x}_0$ using a single sample of Gaussian noise $\boldsymbol{\epsilon} \sim \mathcal{N}(\mathbf{0}, \mathbf{I})$:
\begin{equation}
\mathbf{x}_t = \sqrt{\bar{\alpha}_t}\mathbf{x}_0 + \sqrt{1-\bar{\alpha}_t}\boldsymbol{\epsilon}
\label{eq:forward}
\end{equation}
Here, $\bar{\alpha}_t = \prod_{s=1}^t (1-\beta_s)$, providing a useful abstraction over the original Markov chain.
In fact, both the training objective and the noise scheduler are eventually expressed (and implemented) in these terms.

\paragraph{Reverse Process}
The diffusion model is trained to perform the reverse process $p_\theta(\mathbf{x}_{t-1}|\mathbf{x}_t)$, learning to denoise the data step by step.
There are several ways to do so; in this work, we follow the approach of \citet{ho2020denoising} and model the Gaussian noise $\boldsymbol{\epsilon}$ in Equation~\ref{eq:forward} as a proxy for the cumulative noise at step $t$.
Specifically, a model $\epsilon_\theta(\cdot)$ with parameters $\theta$ is trained to estimate the noise $\boldsymbol{\epsilon}$ given the noised data $\mathbf{x}_t$ and timestep $t$.
In practice, the model often conditions on additional contextual information $c$, such as a caption when generating an image.
The parameters of the noise prediction model are thus optimized by minimizing the mean squared error loss:
\begin{equation}
\mathcal{L}_{\text{DDPM}} = \mathbb{E}_{\mathbf{x}_0, t, \boldsymbol{\epsilon}}\big[||\boldsymbol{\epsilon} - \boldsymbol{\epsilon}_{\theta}(\mathbf{x}_t, t, c)||^2\big]
\end{equation}

\paragraph{Noise Schedule}
When creating a noised example $\mathbf{x}_t$ (Equation~\ref{eq:forward}), $\bar{\alpha}_t$ determines the variance of the noise for timestep $t$.
In this work, we adopt the commonly used cosine scheduler~\cite{nichol2021improved}, which largely follows $\sqrt{\bar{\alpha}_t} \approx \cos(\frac{t}{T}\cdot\frac{\pi}{2})$ with some adjustments.

\paragraph{Inference}
Decoding is done iteratively, pealing away some of the noise at each step.
Starting with pure Gaussian noise at $\mathbf{x}_T$, the model $\boldsymbol{\epsilon}_{\theta}(\mathbf{x}_t, t, c)$ predicts the noise accumulated at timestep $t$.
The predicted noise is then scaled according to the noise schedule, and the proportional amount of predicted noise is removed from $\mathbf{x}_t$ to produce $\mathbf{x}_{t-1}$.
In practice, inference is done over fewer timesteps than training.
Classifier-free guidance (CFG)~\citep{ho2022classifier} is often used to improve generation by contrasting the prediction of the model conditioned on the context $c$ with the unconditioned prediction, at the cost of doubling the computation.

\subsection{Latent Image Representation}
\label{sec:vae}

Early diffusion models worked directly in pixel space \citep{ho2020denoising}, but this proved computationally expensive.
Variational autoencoders (VAEs) \citep{kingma2013auto} can save compute by encoding images into a lower-dimensional latent space. 
Implemented as deep CNNs, modern VAEs are trained on a combination of reconstruction and regularization losses \citep{esser2021taming}, allowing downstream models like latent diffusion models (LDMs) \citep{ldm} to operate efficiently on compact image patch embeddings; e.g. represent every 8$\times$8 pixel patch as an 8-dimensional vector.
For autoregressive language modeling approaches \citep{dalle, yu2022scaling}, images must be discretized.
Discrete autoencoders, such as vector-quantized VAEs (VQ-VAE) \citep{vqvae}, achieve this by introducing a quantization layer (and related regularization losses) that maps continuous latent embeddings to discrete tokens.

\section{Transfusion}
\label{sec:method}

\begin{figure}[t]
  \centering
  \begin{minipage}[b]{0.47\textwidth}
    \centering
    \includegraphics[width=\textwidth]{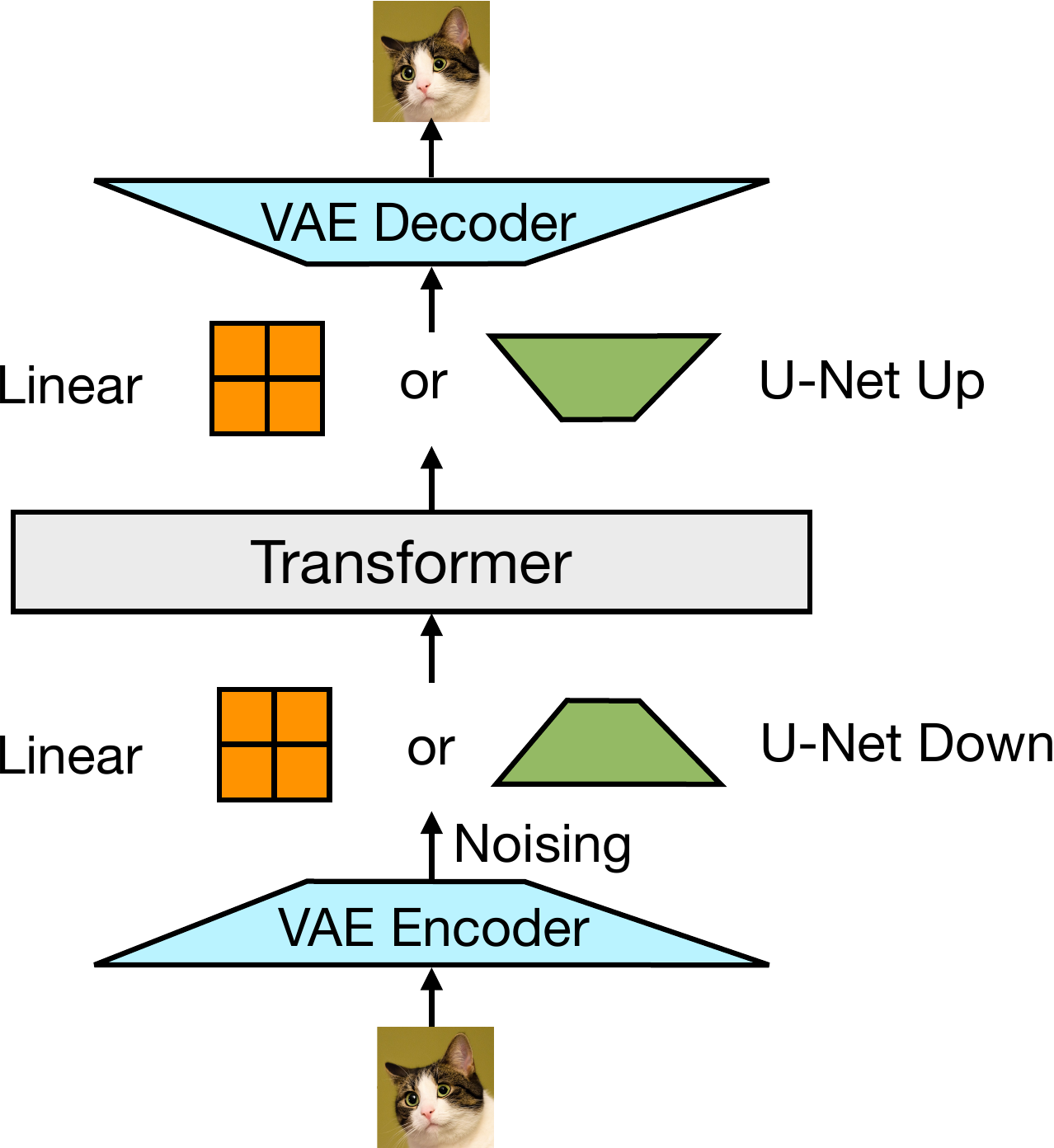}
    \caption{We convert images to and from latent representations using a pretrained VAE, and then into patch representations with either a simple linear layer or U-Net down blocks.}
    \label{fig:encdec}
  \end{minipage}
  \hfill
  \begin{minipage}[b]{0.47\textwidth}
    \centering
    \includegraphics[width=\textwidth]{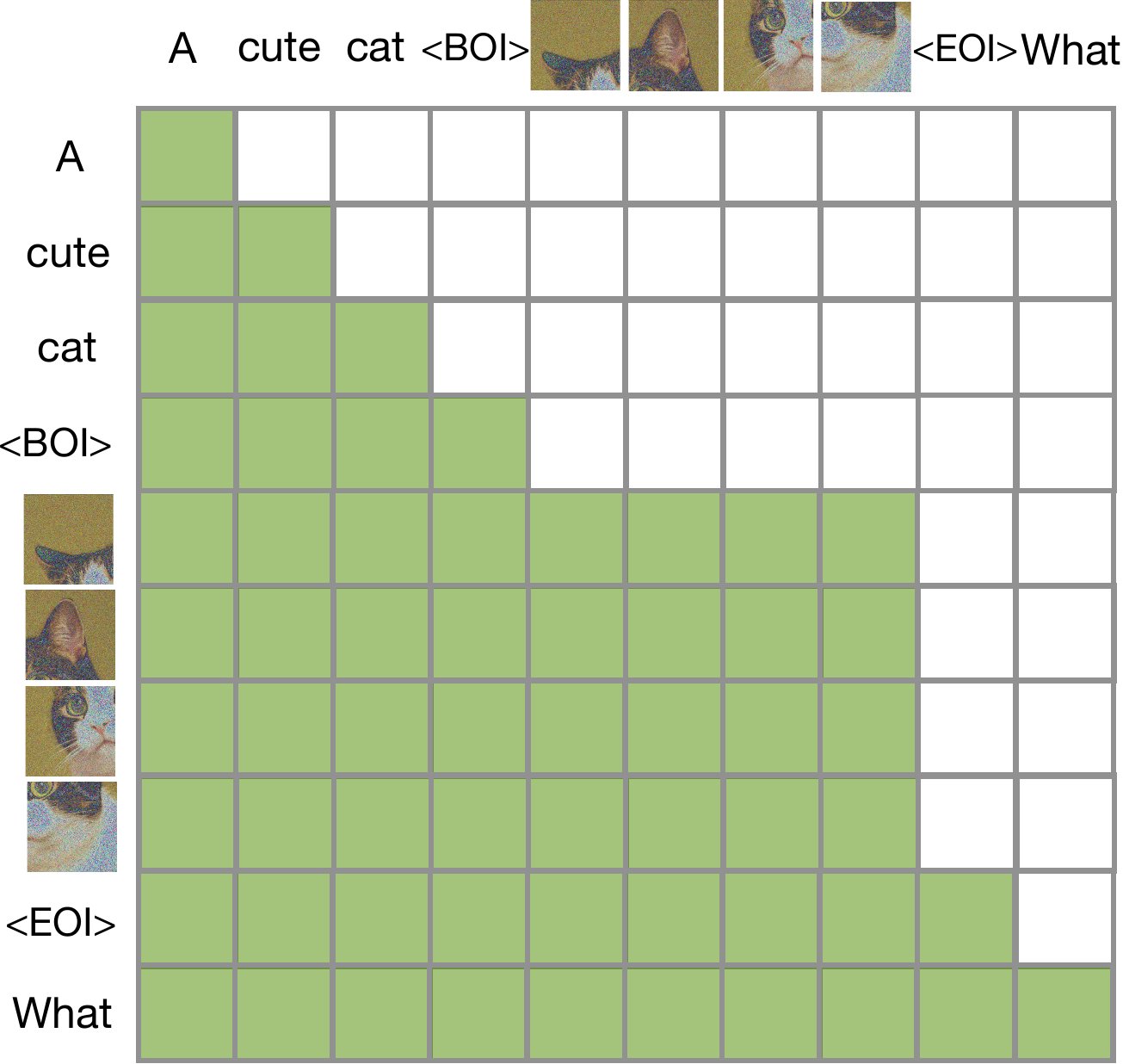}
    \caption{Expanding on the causal mask, Transfusion allows patches of the same image to condition on each other.}
    \label{fig:attention}
  \end{minipage}
\end{figure}

Transfusion is a method for training a single unified model to understand and generate both discrete and continuous modalities.
Our main innovation is demonstrating that we can use separate losses for different modalities -- language modeling for text, diffusion for images -- over shared data and parameters.
Figure~\ref{fig:transfusion} illustrates Transfusion.

\paragraph{Data Representation}
We experiment with data spanning two modalities: discrete text and continuous images.
Each text string is tokenized into a sequence of discrete tokens from a fixed vocabulary, where each token is represented as an integer.
Each image is encoded as latent patches using a VAE (see \S\ref{sec:vae}), where each patch is represented as a continuous vector; the patches are sequenced left-to-right top-to-bottom to create a sequence of patch vectors from each image.\footnote{While our canonical setting uses a VAE following latent diffusion models, we were also able to demonstrate Transfusion using raw pixel representations in preliminary experiments.}
For mixed-modal examples, we surround each image sequence with special \textit{beginning of image} (BOI) and \textit{end of image} (EOI) tokens before inserting it to the text sequence;
thus, we arrive at a single sequence potentially containing both discrete elements (integers representing text tokens) and continuous elements (vectors representing image patches).

\paragraph{Model Architecture}
The vast majority of the model's parameters belong to a single transformer, which processes every sequence, regardless of modality.\footnote{We follow Llama's \citep{touvron2023llama} flavor of the transformer block, which includes the SwiGLU activation function~\citep{shazeer2020glu} and RoPE~\citep{su2024roformer}.}\footnote{While we use the transformer architecture in this work, Transfusion could potentially work with other architectures too, despite its name.}
The transformer takes a sequence of high-dimensional vectors in $\mathbb{R}^d$ as input, and produces similar vectors as output.
To convert our data into this space, we use lightweight modality-specific components with unshared parameters.
For text, these are the embedding matrices, converting each input integer to vector space and each output vector into a discrete distribution over the vocabulary.
For images, we experiment with two alternatives for compressing local windows of $k \times k$ patch vectors into a single transformer vector (and vice versa): (1) a simple linear layer,\footnote{We add an embedding of the timestep $t$ to every patch vector before the linear layer.} and (2) up and down blocks of a U-Net \citep{nichol2021improved,saharia2022photorealistic}.\footnote{We replace the U-Net's AdaLayerNorm with regular layer norm in our implementation.}
Figure~\ref{fig:encdec} illustrates the overall architecture.

\paragraph{Transfusion Attention}
Language models typically use causal masking to efficiently compute the loss and gradients over an entire sequence in a single forward-backward pass without leaking information from future tokens.
While text is naturally sequential, images are not, and are usually modeled with unrestricted (bidirectional) attention.
Transfusion combines both attention patterns by applying causal attention to every element in the sequence, and bidirectional attention within the elements of each individual image.
This allows every image patch to attend to every other patch within the same image, but only attend to text or patches of other images that appeared previously in the sequence.
We find that enabling intra-image attention significantly boosts model performance (see \S\ref{sec:ablations}).
Figure~\ref{fig:attention} shows an example Transfusion attention mask.

\paragraph{Training Objective}
To train our model, we apply the language modeling objective $\mathcal{L}_\text{LM}$ to predictions of text tokens and the diffusion objective $\mathcal{L}_\text{DDPM}$ to predictions of image patches.
LM loss is computed per token,\footnote{When the input is a BOI token, we do not compute any loss.} while diffusion loss is computed per image, which may span multiple elements (image patches) in the sequence.
Specifically, we add noise $\boldsymbol{\epsilon}$ to each input latent image $\mathbf{x}_0$ according to the diffusion process to produce $\mathbf{x}_t$ before patchification, and then compute the image-level diffusion loss.\footnote{Ergo, downstream tokens condition on noisy images during training. See \S\ref{sec:noise} for further discussion.}
We combine the two losses by simply adding the losses computed over each modality with a balancing coefficient $\lambda$:
\begin{equation}
\mathcal{L}_\text{Transfusion} = \mathcal{L}_\text{LM} + \lambda \cdot \mathcal{L}_\text{DDPM}
\label{eq:transfusion}
\end{equation}

This formulation is a specific instantiation of a broader idea: combining a discrete distribution loss with a continuous distribution loss to optimize the same model.
We leave further exploration of this space, such as replacing diffusion with flow matching \citep{lipman2022flow}), to future work.

\paragraph{Inference}
Reflecting the training objective, our decoding algorithm also switches between two modes: LM and diffusion.
In \textit{LM mode}, we follow the standard practice of sampling token by token from the predicted distribution.
When we sample a BOI token, the decoding algorithm switches to \textit{diffusion mode}, where we follow the standard procedure of decoding from diffusion models.
Specifically, we append a pure noise $\mathbf{x}_T$ in the form of $n$ image patches to the input sequence (depending on the desired image size), and denoise over $T$ steps.
At each step $t$, we take the noise prediction and use it to produce $\mathbf{x}_{t-1}$, which then overwrites $\mathbf{x}_{t}$ in the sequence; i.e. the model always conditions on the last timestep of the noised image and cannot attend to previous timesteps.
Once the diffusion process has ended, we append an EOI token to the predicted image, and switch back to LM mode.
This algorithm enables the generation of any mixture of text and image modalities.

\section{Experiments}

We demonstrate in a series of controlled experiments that Transfusion is a viable, scalable method for training a unified multi-modal model.

\subsection{Setup}
\label{sec:setup}

\paragraph{Evaluation}
We evaluate model performance on a collection of standard uni-modal and cross-modal benchmarks (Table~\ref{tab:benchmarks}).
For text-to-text, we measure perplexity on 20M held-out tokens from Wikipedia and the C4 corpus \citep{t5}, as well as accuracy on the pretraining evaluation suite of Llama 2 \citep{llama2}.\footnote{The Llama 2 evaluation suite includes HellaSwag~\citep{zellers2019hellaswag}, PIQA~\citep{bisk2020piqa}, SIQA~\citep{sap2019socialiqa}, WinoGrande~\citep{sakaguchi2021winogrande}, ARC-e and -c~\citep{clark2018think}, and BoolQ~\citep{clark2019boolq}. We report the average 0-shot task accuracy on these benchmarks.}
For text-to-image, we use the MS-COCO benchmark \citep{Eval_mscoco}, where we generate images on randomly selected 30k prompts from validation set and measure their photo-realism using zero-shot Frechet Inception Distance (FID) \citep{fid} as well as their alignment with the prompts using CLIP score \citep{clip}.\footnote{We follow common practice for ablations and use only 5k examples to compute FID and CLIP in \S\ref{sec:ablations}.}
We also evaluate the model's ability to generate image captions; we report CIDEr \citep{cider} scores on the Karpathy test split of MS-COCO \citep{Eval_mscoco}.
These evaluations provide signal for investigation scaling laws (\S\ref{sec:simple}) and ablations (\S\ref{sec:ablations}).
To compare with recent literature in diffusion models, we evaluate our largest scale model (\S\ref{sec:enhanced}) also on GenEval \citep{ghosh2023geneval}, a benchmark that examines a model's ability to generate an accurate depiction of the prompt.

\begin{table}[t]
\centering
\begin{minipage}[b]{0.56\textwidth}
\small
\centering
\begin{tabular}{@{}llll@{}}
\toprule
\textbf{Input} & \textbf{Output} & \textbf{Benchmark} & \textbf{Metric} \\
\midrule
\multirow{3}{*}{Text} & \multirow{3}{*}{Text} & Wikipedia & Perplexity ($\downarrow$) \\
& & C4 & Perplexity ($\downarrow$) \\
& & Llama 2 Eval Suite & Accuracy ($\uparrow$) \\
\midrule
Image & Text & MS-COCO 5k & CIDEr ($\uparrow$) \\
\midrule
\multirow{2}{*}{Text} & \multirow{2}{*}{Image} & MS-COCO 30k & FID ($\downarrow$), CLIP ($\uparrow$) \\
& & GenEval & GenEval score ($\uparrow$) \\
\bottomrule \\
\end{tabular}
\caption{An overview of the evaluation suite used in this work.}
\label{tab:benchmarks}
\end{minipage}
\hfill
\begin{minipage}[b]{0.41\textwidth}
\small
\centering
\begin{tabular}{@{}lrrr@{}}
\toprule
\textbf{Size} & \textbf{Layers} & \textbf{Emb Dim} & \textbf{Att Heads} \\
\midrule
0.16B & 16 & 768  & 12 \\
0.37B & 24 & 1024 & 16 \\
0.76B & 24 & 1536 & 24 \\
1.4B  & 24 & 2048 & 16 \\
7B    & 32 & 4096 & 32 \\
\bottomrule \\
\end{tabular}
\caption{Model sizes and configurations for both Transfusion and baselines.}
\label{tab:models}
\end{minipage}
\end{table}

\paragraph{Baseline}
At the time of writing, the prominent open-science method for training a single mixed-modal model that can generate both text and images is to quantize images into discrete tokens, and then model the entire token sequence with a standard language model \citep{dalle, yu2022scaling, yu2023scaling}.
We follow the recipe of Chameleon \citep{team2024chameleon} to train a family of data- and compute-controlled baseline models, which we can directly compare to our Transfusion models.
The key difference between Chameleon and Transfusion is that while Chameleon discretizes images and processes them as tokens, Transfusion keeps images in continuous space, removing the quantization information bottleneck.
To further minimize any confounding variables, we train the VAEs for Chameleon and Transfusion using exactly the same data, compute, and architecture, with the only differentiator being the quantization layer and codebook loss of Chameleon's VQ-VAE (see details below).
Chameleon also deviates from the Llama transformer architecture, adding query-key normalization, post-normalization, denominator loss, and a lower learning rate of 1e-4 to manage training instability, which incur an efficiency cost (see \S\ref{sec:simple}).\footnote{Removing these deviations in preliminary experiments encountered optimization instabilities in Chameleon.}

\paragraph{Data}
For almost all of our experiments, we sample 0.5T tokens (patches) from two datasets at a 1:1 token ratio.
For text, we use the Llama 2 tokenizer and corpus \citep{llama2}, containing 2T tokens across a diverse distribution of domains.
For images, we use a collection of 380M licensed Shutterstock images and captions.
Each image is center-cropped and resized to produce a 256$\times$256 pixel image.\footnote{Depending on the compression rate of the patch encoder (see Model Architecture in \S\ref{sec:method}), each image will be represented by either 1024, 256, 64, or 16 elements in the sequence. Since the text/image ratio is constant during training, higher compression rates enable training on more images in total, at the cost of less compute per image.}
We randomly order the image and captions, ordering the caption first 80\% of the time.

In one experiment (\ref{sec:enhanced}) we scale up the total training data to 2T tokens (1T text tokens and about 3.5B caption-image pairs at 256 patches per image).
To diversify, we add 220M publicly available images with captions, prefiltered to not contain people.
To rebalance the distribution, we upsample 80M Shutterstock images containing people.
We also add data from Conceptual 12M (CC12M) \citep{cc12m}, reaching a total mixture of 692M image-caption pairs per epoch.
Finally, we upweight the portion of high-aesthetic images in the last 1\% of the training schedule.

\paragraph{Latent Image Representation}
We train a 86M parameter VAE following \citet{esser2021taming}.
We use a CNN encoder and decoder, and latent dimension 8.
The training objective is combines reconstruction and regularization losses.\footnote{See Appendix~\ref{sec:vae_details} for details.}
Our implementation reduces an image of 256$\times$256 pixels to a 32$\times$32$\times $8 tensor, where each latent 8-dimensional latent pixel represents (conceptually) an 8$\times$8 pixel patch in the original image, and trains for 1M steps.
For VQ-VAE training, we follow the same setup described for VAE training, except we replace $\mathcal{L}_{\text{KL}}$ with the standard codebook commitment loss with $\beta = 0.25$ \citep{vqvae}. We use a codebook of 16,384 token types.

\paragraph{Model Configuration}
To investigate scaling trends, we train models at five different sizes -- 0.16B, 0.37B, 0.76B, 1.4B, and 7B parameters -- following the standard settings from Llama \citep{touvron2023llama}.
Table~\ref{tab:models} describes each setting in detail.
In configurations that use linear patch encoding (\S\ref{sec:simple} and \S\ref{sec:ablations}), the number of additional parameters is insignificant, accounting for fewer than 0.5\% of total parameters in every configuration.
When using U-Net patch encoding (\S\ref{sec:ablations} and \S\ref{sec:enhanced}), these parameters add up to 0.27B additional parameters across all configurations; while this is a substantial addition of parameters to smaller models, these layers amount to only a 3.8\% increase of the 7B configuration, almost identical to the number of parameters in the embedding layers.

\paragraph{Optimization}
We randomly initialize all model parameters, and optimize them using AdamW ($\beta_1=$0.9, $\beta_2=$0.95, $\epsilon=$1e-8) with a learning rate of 3e-4, warmed up for 4000 steps and decaying to 1.5e-5 using a cosine scheduler.
We train on sequences of 4096 tokens in batches of 2M tokens for 250k steps, reaching 0.5T tokens in total.
In our large-scale experiment (\S\ref{sec:enhanced}), we train with a batch size of 4M tokens over 500k steps, totalling 2T tokens.
We regularize with weight decay of 0.1 and clip gradients by norm (1.0).
We set the $\lambda$ coefficient in the Transfusion objective (Equation~\ref{eq:transfusion}) to 5 following preliminary experiments;
we leave further tuning of $\lambda$ to future work.

\paragraph{Inference}
In text mode, we use greedy decoding for generating text.
Ranked classification is used for the Llama evaluation suite.
For image generation, we follow the standard of 250 diffusion steps (the model is trained on 1,000 timesteps).
We follow Chameleon and use CFG with a coefficient of 5 in the controlled comparison experiments (\S\ref{sec:simple}).
This value is suboptimal for Transfusion, and so we use a CFG coefficient of 3 throughout the ablation experiments (\S\ref{sec:ablations}), and follow the standard practice of tuning the coefficient for each benchmark in our large scale experiment (\S\ref{sec:enhanced}).

\subsection{Controlled Comparison with Chameleon}
\label{sec:simple}

We run a series of controlled experiments to compare Transfusion with Chameleon at different model sizes ($N$) and token counts ($D$), using the combination of both as a proxy for FLOPs ($6ND$).\footnote{Since Transfusion uses continuous representations of images, it can express a single image with significantly fewer tokens, shortening the average document length and thus the overall quadratic price of attention. Since this fact favors Transfusion, we remove this confounder by using the theoretical FLOP calculation.}
For simplicity and parameter control, the Transfusion variant in these experiments uses simple linear image encoder/decoder with patch size 2$\times$2, as well as bidirectional attention.
For each benchmark, we plot all results on a log-metric over log-FLOPs curve and regress linear trendlines.\footnote{The smaller Chameleon models perform poorly on image generation and understanding tasks, leading to outlier results that do not correlate with the emerging scaling law of larger Chameleon models. We therefore define minimal performance thresholds below which we remove datapoints: $\leq$100 FID, $\geq$17 CLIP, $\geq$4 CIDEr.}
We also estimate relative compute efficiency by measuring the parity FLOP ratio: the ratio between the number of FLOPs required by Transfusion and Chameleon to reach the same level of performance.

\begin{figure}[!t]
  \centering
  \begin{subfigure}[b]{0.49\textwidth}
    \centering
    \includegraphics[width=\textwidth]{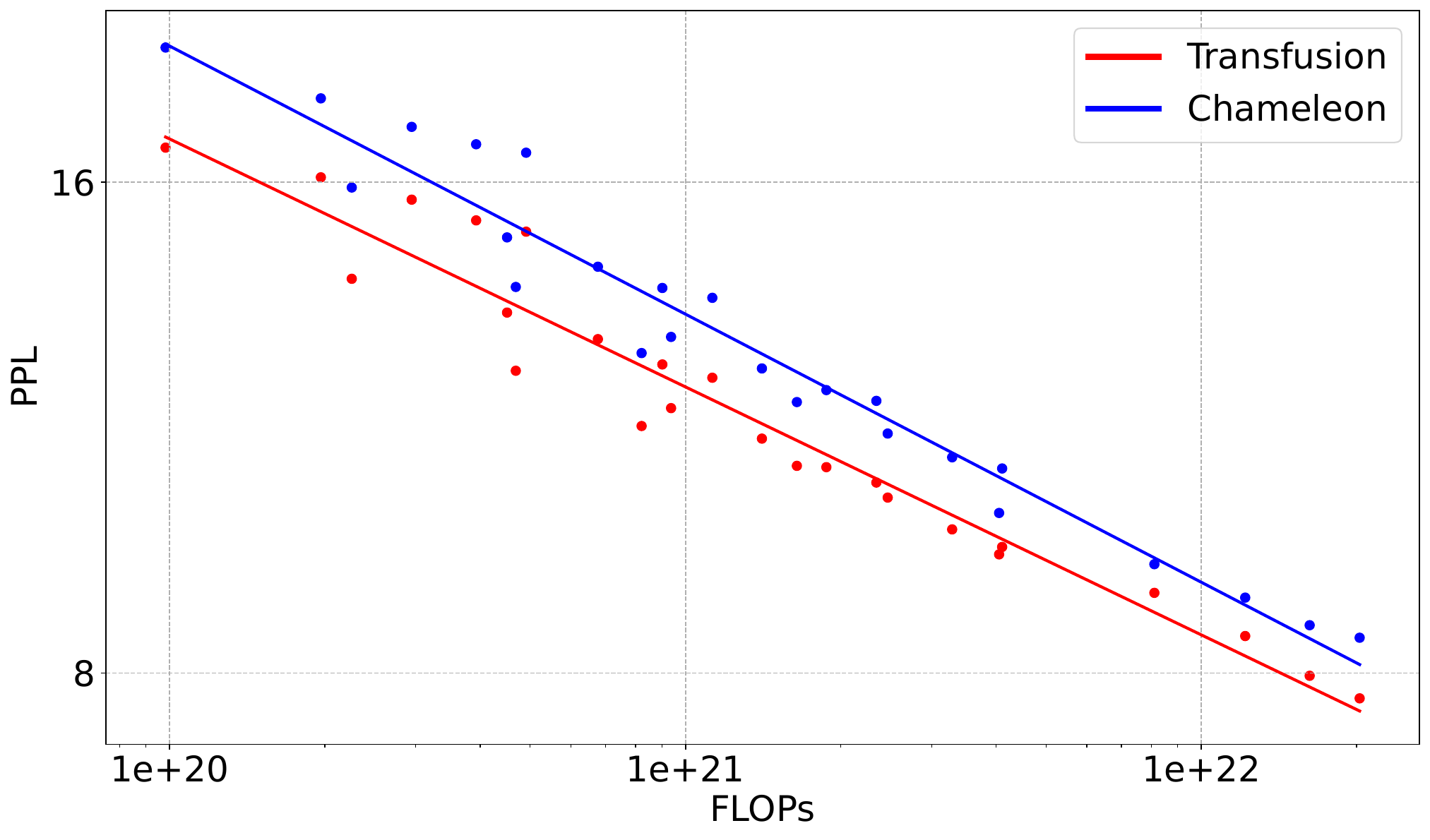}
    \caption{C4 Perplexity}
  \end{subfigure}
  \hfill
  \begin{subfigure}[b]{0.49\textwidth}
    \centering
    \includegraphics[width=\textwidth]{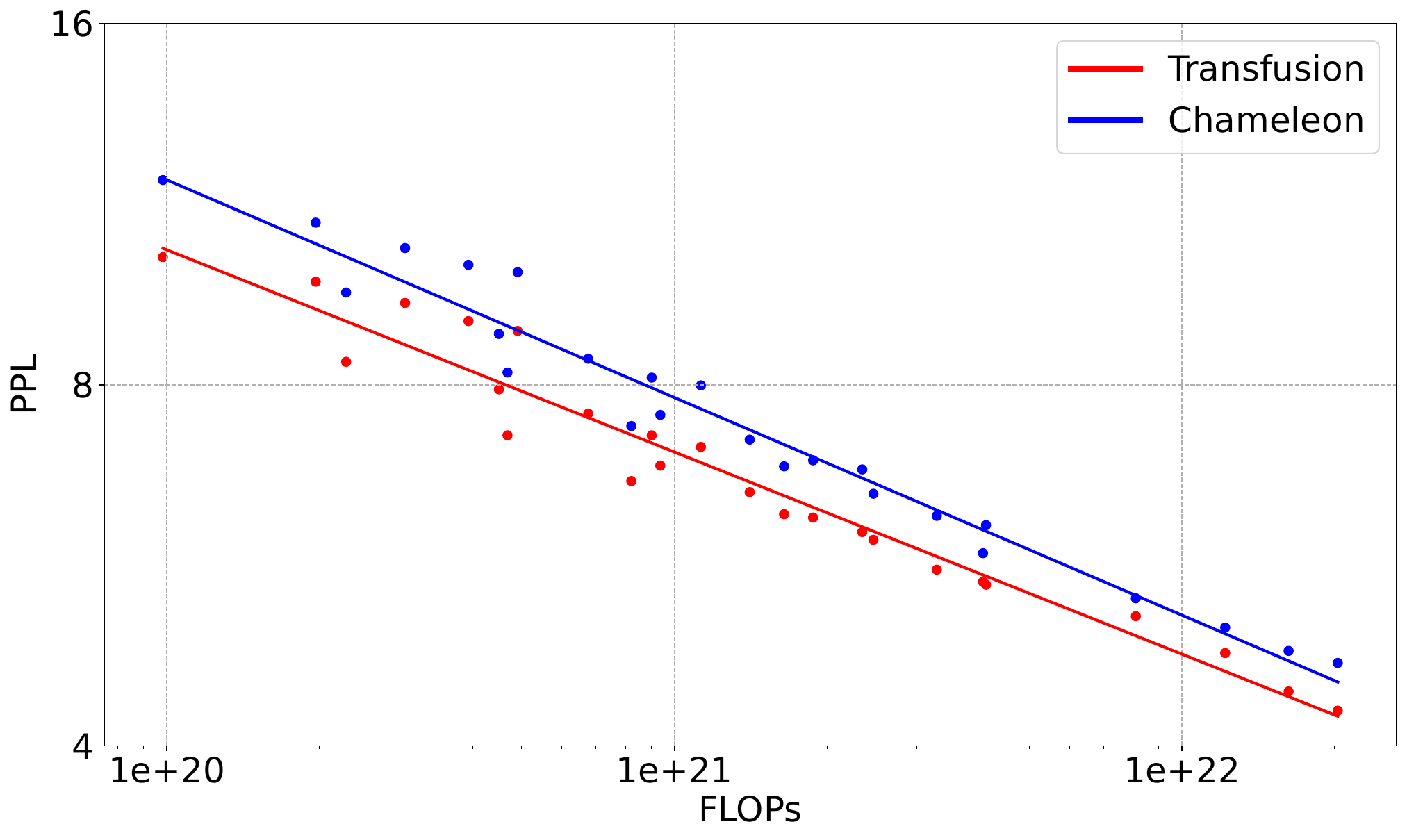}
    \caption{Wikipedia Perplexity}
  \end{subfigure}

  \vspace{1em}

  \begin{subfigure}[b]{0.49\textwidth}
    \centering
    \includegraphics[width=\textwidth]{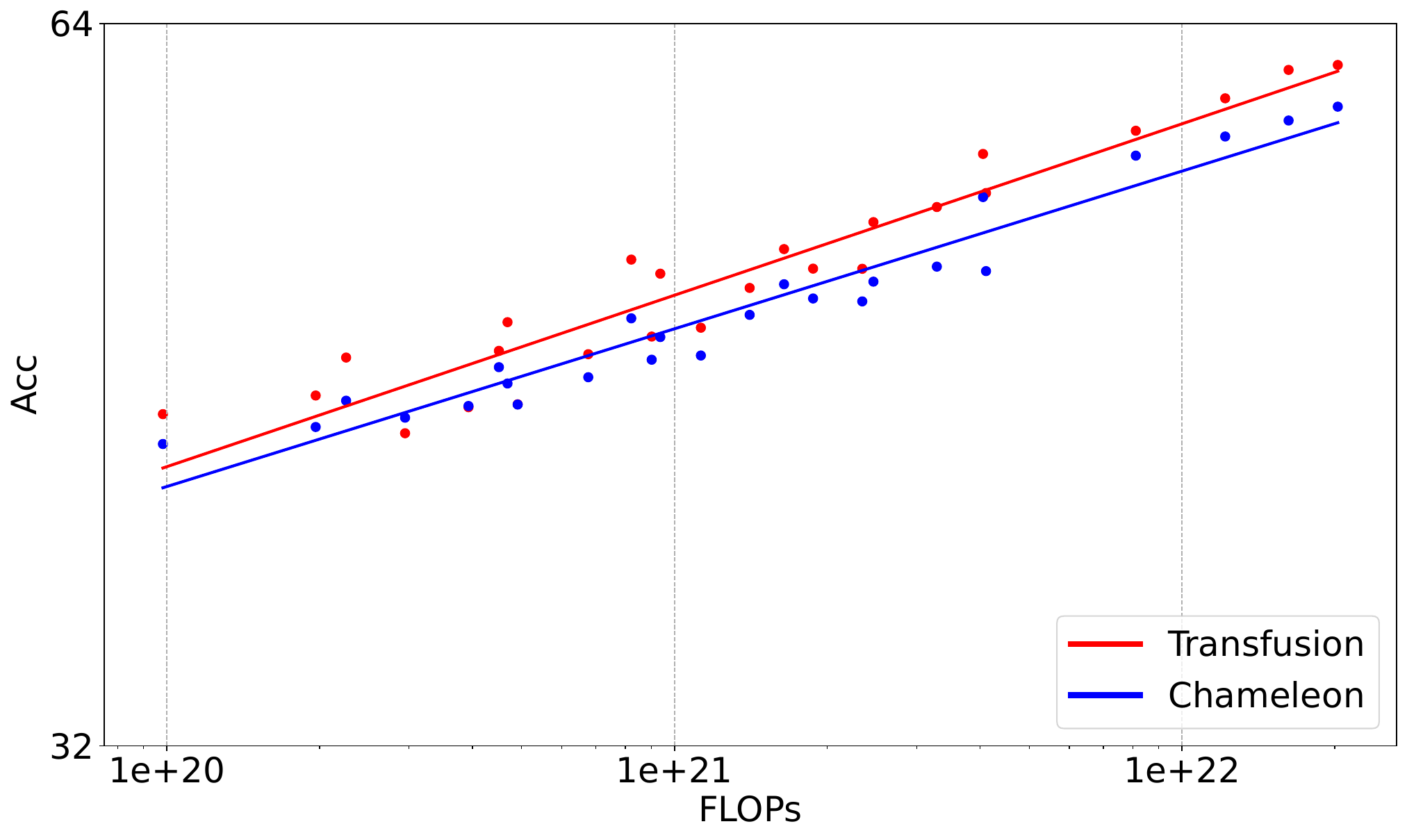}
    \caption{Llama 2 Eval Suite Accuracy}
  \end{subfigure}
  \hfill
  \begin{subfigure}[b]{0.49\textwidth}
    \centering
    \includegraphics[width=\textwidth]{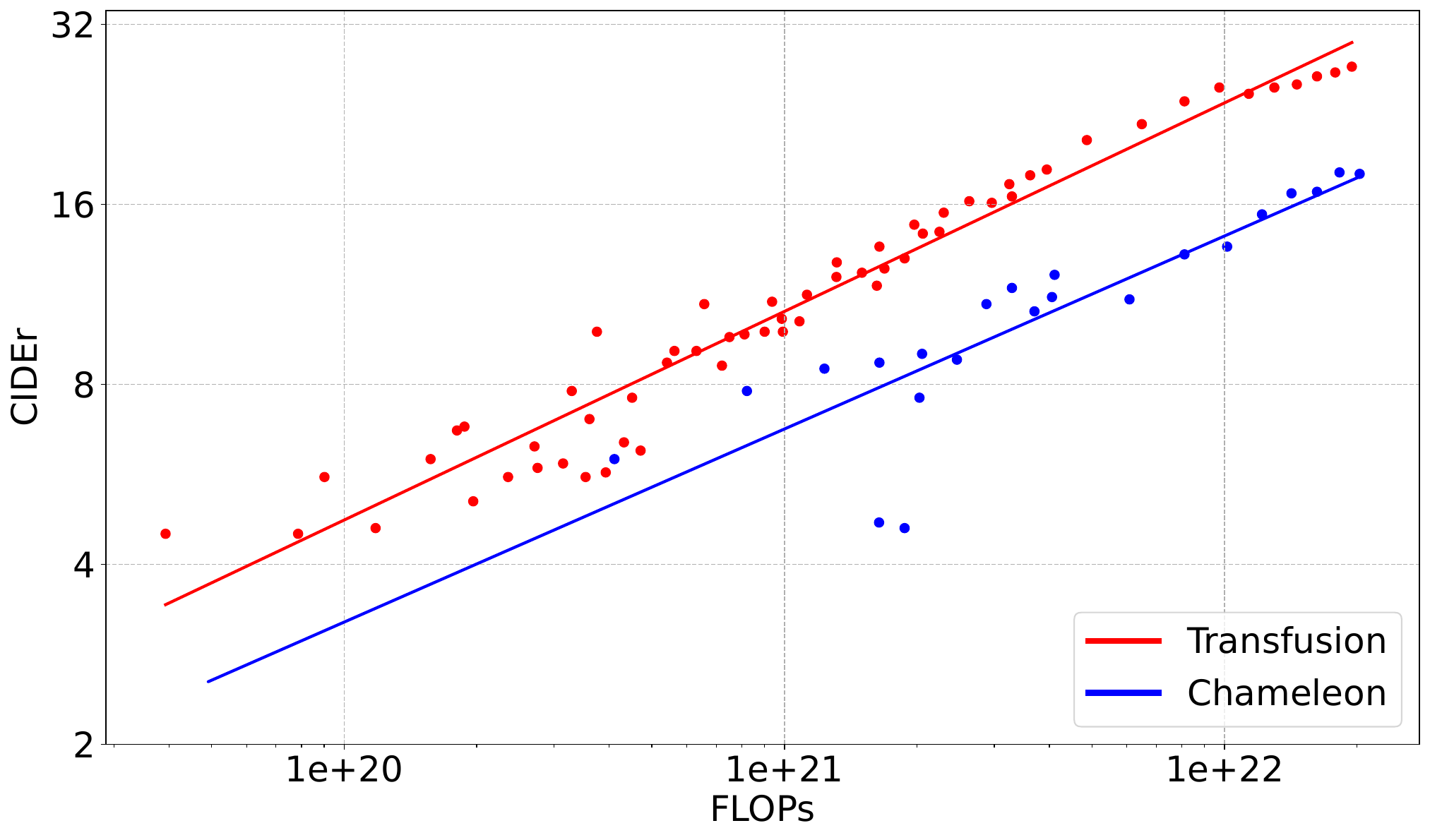}
    \caption{MS-COCO 5k CIDEr}
  \end{subfigure}
  
  \vspace{1em}

  \begin{subfigure}[b]{0.49\textwidth}
    \centering
    \includegraphics[width=\textwidth]{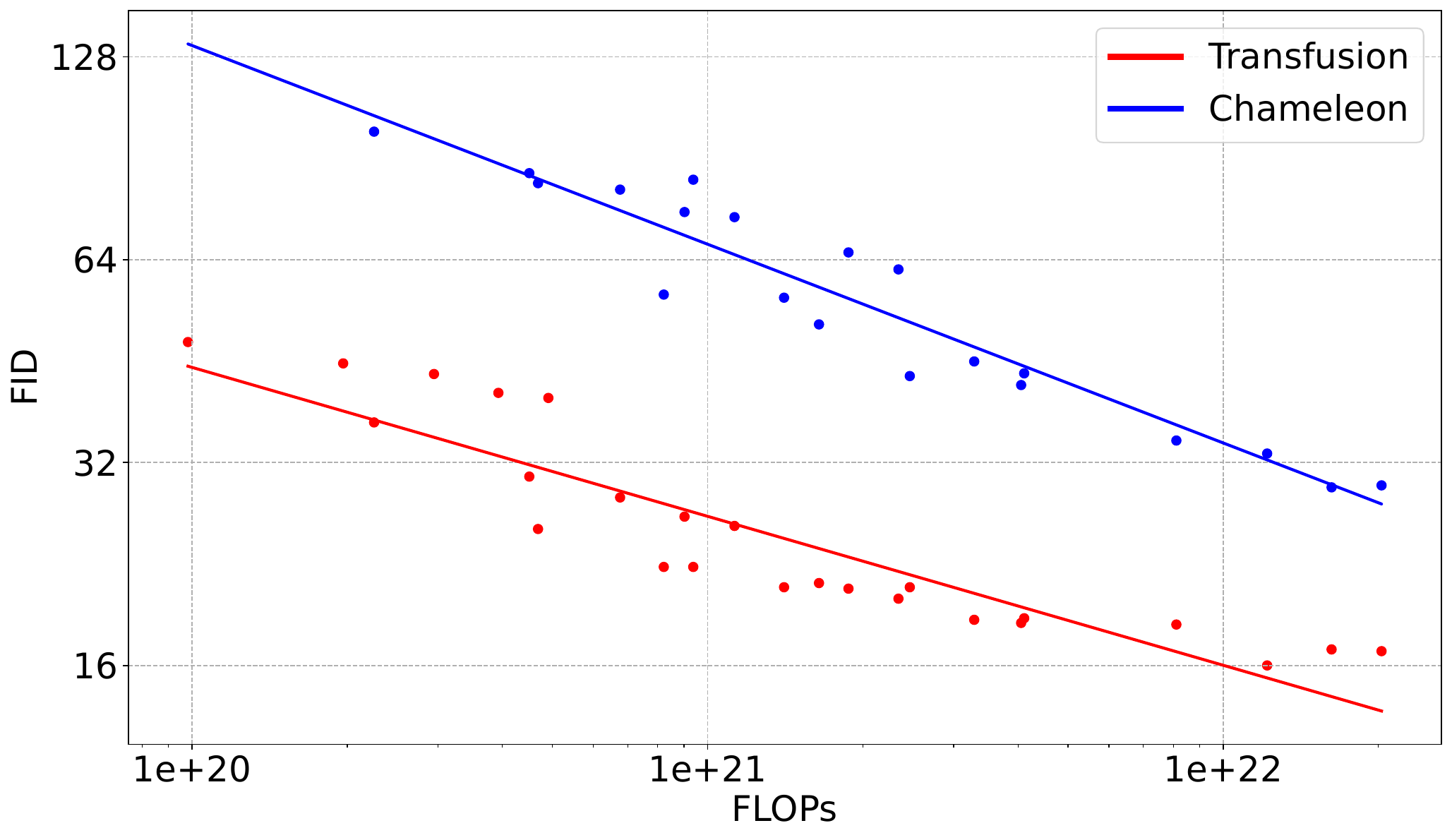}
    \caption{MS-COCO 30k FID}
  \end{subfigure}
  \hfill
  \begin{subfigure}[b]{0.49\textwidth}
    \centering
    \includegraphics[width=\textwidth]{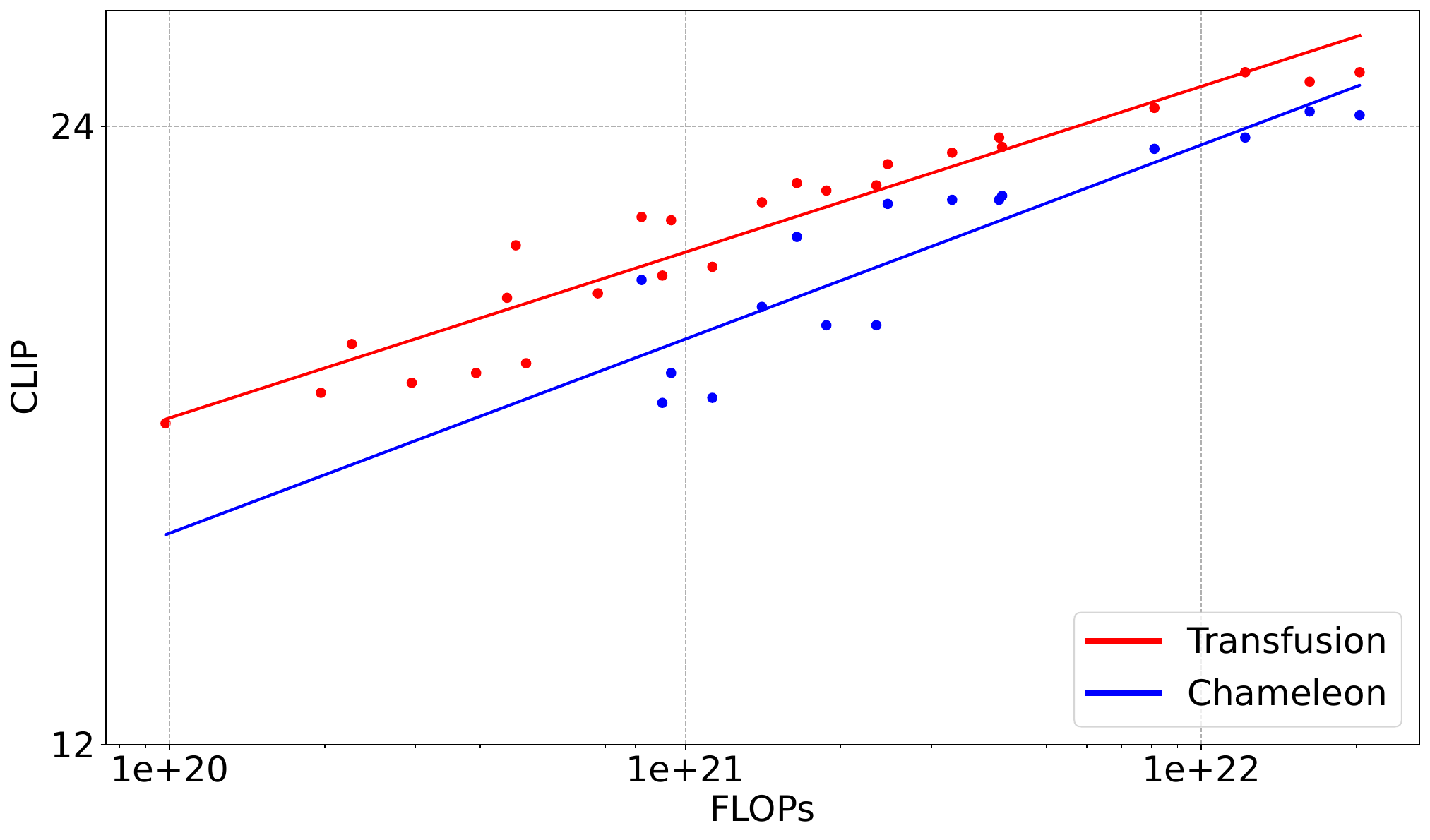}
    \caption{MS-COCO 30k CLIP}
  \end{subfigure}
  \caption{Performance of Transfusion and Chameleon models at different scales, controlled for parameters, data, and compute. All axes are logarithmic.}
  \label{fig:controlled}
\end{figure}

Figure~\ref{fig:controlled} visualizes the scaling trends,
and Table~\ref{tab:controlled} shows the results of the largest models in this controlled setting and their estimated parity FLOP ratio.
In every benchmark, Transfusion consistently exhibits better scaling laws than Chameleon.
While the lines are close to parallel, there is a significant gap in Transfusion's favor.
The difference in compute efficiency is particularly striking in image generation, where FID Transfusion achieves parity with Chameleon using 34$\times$ less compute.

Surprisingly, text-only benchmarks also reveal better performance with Transfusion, even though both Transfusion and Chameleon model text in the same way.
We investigate this phenomenon by ablating the various changes leading up to Transfusion and Chameleon from the original Llama 2 recipe.
Table~\ref{tab:text_analysis} shows that while Transfusion does come at a non-zero cost to text performance, the Chameleon recipe suffers from both the stability modifications made to the architecture and from the introduction of image tokens.
Training on quantized image tokens degrades text performance more than diffusion on all three benchmarks.
One hypothesis is that this stems from the competition between text and image tokens in the output distribution;
alternatively, it is possible that diffusion is more efficient at image generation and requires fewer parameters, allowing Transfusion models to use more capacity than Chameleon to model text.
We leave further investigation of this phenomenon to future research.

\vfill

\begin{table}[!b]
\small
\centering
\begin{tabular}{@{}lrrrrrr@{}}
\toprule
\multirow{2}{*}{\textbf{Model}} & \multicolumn{1}{c}{\textbf{C4}} & \multicolumn{1}{c}{\textbf{Wiki}} & \multicolumn{1}{c}{\textbf{Llama}} & \multicolumn{3}{c}{\textbf{MS-COCO}} \\
& \textbf{PPL} ($\downarrow$) & \textbf{PPL} ($\downarrow$) & \textbf{Acc} ($\uparrow$) & \textbf{CDr} ($\uparrow$) & \textbf{FID} ($\downarrow$) & \textbf{CLIP} ($\uparrow$) \\
\midrule
Transfusion & \textbf{7.72} & \textbf{4.28} & \textbf{61.5}  & \textbf{27.2}  & \textbf{16.8} & \textbf{25.5} \\
Chameleon   & 8.41 & 4.69 & 59.1  & 18.0    & 29.6   & 24.3 \\
\midrule
Parity FLOP Ratio & 0.489 & 0.526 & 0.600 & 0.218 & 0.029 & 0.319 \\
\bottomrule \\
\end{tabular}
\caption{Performance of the largest (7B) Transfusion and Chameleon models in a controlled setting. Both models were trained on 0.5T tokens. \textbf{Parity FLOP Ratio} is the relative amount of Transfusion FLOPs needed to match the results of Chameleon 7B.}
\label{tab:controlled}
\end{table}

\begin{table}[t]
\small
\centering
\begin{tabular}{@{}lllrrr@{}}
\toprule
\multirow{2}{*}{\textbf{Model}} & & \multirow{2}{*}{\textbf{Batch}} & \multicolumn{1}{c}{\textbf{C4}} & \multicolumn{1}{c}{\textbf{Wiki}} & \textbf{Llama} \\
& & & \textbf{PPL} ($\downarrow$)   & \textbf{PPL} ($\downarrow$)  & \textbf{Acc} ($\uparrow$) \\
\midrule
Llama 2      &                                                   & 1M Text Tokens            & 10.1 & 5.8  & 53.7 \\
\midrule
Transfusion & + Diffusion            & + 1M Image Patches & (+0.3) 10.4 & (+0.2) 6.0 & (-2.0) 51.7 \\
\midrule
Chameleon   & + Stability Modifications & 1M Text Tokens            & (+0.9) 11.0 & (+0.5) 6.3 & (-1.8) 51.9 \\
            & + LM Loss on Image Tokens & + 1M Image Tokens & (+0.8) 11.8 & (+0.5) 6.8  & (-3.0) 48.9 \\
\bottomrule \\
\end{tabular}
\caption{Performance of the 0.76B Transfusion and Chameleon models on text-only benchmarks, compared to the original Llama 2 recipe.}
\label{tab:text_analysis}
\end{table}

\subsection{Architecture Ablations}
\label{sec:ablations}

Now that we have established that Transfusion is a viable, scalable approach to multi-modal modeling in a controlled environment, we can explore improvements and extensions that are applicable to Transfusion alone.

\subsubsection{Attention Masking}
\label{sec:attention}

We first examine the necessity of intra-image bidirectional attention.
Table~\ref{tab:attention} shows that enabling this attention pattern beyond the standard causal attention is advantageous throughout all benchmarks, and using both image encoding/decoding architectures.
In particular, we notice a significant improvement in FID when using linear encoding layers (61.3$\rightarrow$20.3).
In the causal-only version of this architecture, there is no flow of information from patches that appear later in the sequence to those before;
since U-Net blocks contain bidirectional attention within, independent of the transformer's attention mask, this gap is less pronounced when they are applied.

\begin{table}[t]
\small
\centering
\begin{tabular}{@{}llrrrrrr@{}}
\toprule
\multirow{2}{*}{\textbf{Enc/Dec}} & \multirow{2}{*}{\textbf{Attention}} & \multicolumn{1}{c}{\textbf{C4}} & \multicolumn{1}{c}{\textbf{Wiki}} & \multicolumn{1}{c}{\textbf{Llama}} & \multicolumn{3}{c}{\textbf{MS-COCO}} \\
& & \textbf{PPL} ($\downarrow$) & \textbf{PPL} ($\downarrow$) & \textbf{Acc} ($\downarrow$) & \textbf{CDr} ($\uparrow$) & \textbf{FID} ($\downarrow$) & \textbf{CLIP} ($\uparrow$) \\
\midrule
\multirow{2}{*}{Linear} & Causal & 10.4 & 6.0 &  51.4 & 12.7 & 61.3 & 23.0 \\
& Bidirectional & 10.4 & 6.0 & 51.7 & 16.0 & 20.3 & 24.0 \\
\midrule
\multirow{2}{*}{U-Net} & Causal & 10.3 & 5.9  &  52.0 & 23.3 &16.8 & 25.3 \\
& Bidirectional & 10.3 & 5.9 &  51.9 & 25.4 & 16.7 & 25.4 \\
\bottomrule \\
\end{tabular}
\caption{Performance of 0.76B Transfusion models with and without intra-image bidirectional attention. Patch size is set at 2$\times$2 latent pixels.}
\label{tab:attention}
\end{table}

\subsubsection{Patch Size}
\label{sec:patch}

Transfusion models can be defined over different sizes of latent pixel patches.
Larger patch sizes allow the model to pack more images in each training batch and dramatically reduce inference compute, but may come at a performance cost.
Table~\ref{tab:patch_size} sheds light on these performance trade-offs.
While performance does decrease consistently as each image is represented by fewer patches with linear encoding, models with U-Net encoding benefit from larger patches on tasks involving the image modality.
We posit that this is due to the greater amount of total images (and diffusion noise) seen during training. 
We also observe that text performance deteriorates with larger patches, perhaps because transfusion needs to exert more resources (i.e. parameters) to learn how to process images with fewer patches and thus less inference compute.

\begin{table}[t]
\small
\centering
\begin{tabular}{@{}lcccrrrrrr@{}}
\toprule
\multirow{2}{*}{\textbf{Enc/Dec}} & \textbf{Latent/} & \textbf{Pixel/} & \textbf{Patch/} & \multicolumn{1}{c}{\textbf{C4}} & \multicolumn{1}{c}{\textbf{Wiki}} & \multicolumn{1}{c}{\textbf{Llama}} & \multicolumn{3}{c}{\textbf{MS-COCO}} \\
& \textbf{Patch} & \textbf{Patch} & \textbf{Image} & \textbf{PPL} ($\downarrow$) & \textbf{PPL} ($\downarrow$) & \textbf{Acc} ($\downarrow$) & \textbf{CDr} ($\uparrow$) & \textbf{FID} ($\downarrow$) & \textbf{CLIP} ($\uparrow$) \\                 
\midrule
None & 1$\times$1  & 8$\times$8 & 1024 & \textbf{10.3} & \textbf{5.9}  & \textbf{52.2}  & 12.0    &    21.0         & 24.0   \\
\midrule
\multirow{3}{*}{Linear}  & 2$\times$2 & 16$\times$16 & ~~256 & \underline{10.4} & \underline{6.0}    & \underline{51.7}  & \underline{16.0}    & \underline{20.3}        & \underline{24.0}   \\
                       & 4$\times$4  & 32$\times$32 & ~~~~64 & 10.9 & 6.3  & 49.8  & 14.3  & 25.6        & 22.6 \\
                       & 8$\times$8  & 64$\times$64 & ~~~~16 & 11.7 & 6.9  & 47.7  & 11.3  & 43.5        & 18.9 \\
\midrule
\multirow{3}{*}{U-Net} & 2$\times$2 & 16$\times$16 & ~~256 & \textbf{\underline{10.3}} & \textbf{\underline{5.9}}  & \underline{51.9}  & 25.4  & 16.7        & 25.4 \\
                       & 4$\times$4  & 32$\times$32 & ~~~~64 & 10.7 & 6.2  & 50.7  & \textbf{\underline{29.9}}  & \textbf{\underline{16.0}} & \textbf{\underline{25.7}} \\
                       & 8$\times$8 & 64$\times$64 & ~~~~16
                       & 11.4 & 6.6  & 49.2  & 29.5  & 16.1        & 25.2 \\
\bottomrule \\
\end{tabular}
\caption{Performance of 0.76B Transfusion models with different patch sizes. Bolded figures indicate global best, underlines indicate best within architecture.}
\label{tab:patch_size}
\end{table}

\subsubsection{Patch Encoding/Decoding Architecture}
\label{sec:architecture}

Our experiments so far indicate an advantage to using the U-Net up and down blocks instead of a simple linear layer.
One possible reason is that the model benefits from the inductive biases of the U-Net architecure;
an alternative hypothesis is that this advantage stems from the significant increase in overall model parameters introduced by the U-Net layers.
To decouple these two confounders, we scale up the core transformer to 7B parameters, while keeping the amount of U-Net parameters (almost) constant;\footnote{While we do not scale the U-Net layers with the transformer in these experiments, this is a potentially fruitful avenue for future research.}
in this setting, the additional encoder/decoder parameters account for only a 3.8\% increase of total model parameters, equivalent to the amount of token embedding parameters.

Table~\ref{tab:patch_arch} shows that even though the relative benefit of U-Net layers shrinks as the transformer grows, it does not diminish.
In image generation, for example, the U-Net encoder/decoder allows much smaller models to obtain better FID scores than the 7B model with linear patchification layers.
We observe a similar trend in image captioning, where adding U-Net layers boosts the CIDEr score of a 1.4B transformer (1.67B combined) beyond the performance of the linear 7B model.
Overall, it appears that there are indeed inductive bias benefits to U-Net encoding and decoding of images beyond the mere addition of parameters.

\begin{table}[t]
\small
\centering
\begin{tabular}{@{}llcrrrrrr@{}}
\toprule
\textbf{Model} & \multirow{2}{*}{\textbf{Enc/Dec}} & \textbf{$\Delta$ Enc/Dec} & \multicolumn{1}{c}{\textbf{C4}} & \multicolumn{1}{c}{\textbf{Wiki}} & \multicolumn{1}{c}{\textbf{Llama}} & \multicolumn{3}{c}{\textbf{MS-COCO}} \\
\textbf{Params} &  & \textbf{Params} & \textbf{PPL} ($\downarrow$) & \textbf{PPL} ($\downarrow$) & \textbf{Acc} ($\uparrow$) & \textbf{CDr} ($\uparrow$) & \textbf{FID} ($\downarrow$) & \textbf{CLIP} ($\uparrow$) \\
\midrule
\multirow{2}{*}{0.16B} & Linear & ~~~~0.5\% & 14.8 &	8.8	 &  44.2  &  6.2 & 37.6 & 20.0\\
                       & U-Net  & 106.1\%   & 14.4 &	8.5	 &  45.7  &  15.3 &	18.8 & 23.9\\
\midrule
\multirow{2}{*}{0.37B} & Linear & ~~~~0.4\% & 12.0 &	7.0	 &  47.9  &  11.1 &	21.5 & 22.4 \\
                       & U-Net  & ~~71.3\%  & 11.8 &	 6.9  &  48.8  &  21.1 & 18.1 & 24.9 \\
\midrule
\multirow{2}{*}{0.76B} & Linear & ~~~~0.4\% & 10.4 &	6.0	 &  51.7  &  16.0 &	20.3 & 24.0 \\
                       & U-Net  & ~~35.5\%  & 10.3 &	5.9	 &  51.9  &  25.4 &	16.7 & 25.4\\
\midrule
\multirow{2}{*}{1.4B} & Linear & ~~~~0.4\%  & 9.5	 &  5.4  &	53.8  &  19.1 &	19.4 & 24.3\\
                      & U-Net  & ~~19.3\%   & 9.4  &  5.4  &  53.4  &  28.1 & 16.6 & 25.7\\
\midrule
\multirow{2}{*}{7B}   & Linear & ~~~~0.3\%  & 7.7  &  4.3  &  61.5  &  27.2 &	18.6 & 25.9\\
                      & U-Net  & ~~~~3.8\%  & 7.8  &  4.3  &  61.1  &  33.7 & 16.0 & 26.5\\
\bottomrule \\
\end{tabular}
\caption{Performance of linear and U-Net variants of Transfusion across different model sizes. Patch size is set at 2$\times$2 latent pixels. Model parameters refers to the transformer alone.}
\label{tab:patch_arch}
\end{table}

\subsubsection{Image Noising}
\label{sec:noise}

Our experiments order 80\% of image-caption pairs with the caption first, and the image conditioning on the caption, following the intuition that image generation may be a more data-hungry task than image understanding.
The remaining 20\% of the pairs condition the caption on the image.
However, these images are noised as part of the diffusion objective.
We thus measure the effect of limiting the diffusion noise to a maximum of $t=500$ (half of the noise schedule) in the 20\% of cases where images appear before their captions.
Table~\ref{tab:noise_cap} shows that noise limiting significantly improves image captioning, as measure by CIDEr, while having a relatively small effect (less than 1\%) on other benchmarks.

\begin{table}[t]
\small
\centering
\begin{tabular}{@{}lcrrrrrr@{}}
\toprule
\textbf{Model} & \textbf{Noise} & \multicolumn{1}{c}{\textbf{C4}} & \multicolumn{1}{c}{\textbf{Wiki}} & \multicolumn{1}{c}{\textbf{Llama}} & \multicolumn{3}{c}{\textbf{MS-COCO}} \\
\textbf{Params} & \textbf{Limit} & \textbf{PPL} ($\downarrow$) & \textbf{PPL} ($\downarrow$) & \textbf{Acc} ($\uparrow$) & \textbf{CDr} ($\uparrow$) & \textbf{FID} ($\downarrow$) & \textbf{CLIP} ($\uparrow$) \\
\midrule
\multirow{2}{*}{0.76B} &  & 10.3  & 5.9 &  51.9 &  25.4 & 16.7 & 25.4 \\
 & $\checkmark$ & 10.3  & 5.9 &  52.1 &  \textbf{29.4} & 16.5 & 25.4 \\
\midrule
\multirow{2}{*}{7B}					  &   & 7.8  &  4.3 &  61.1  & 33.7 & 16.0 & 26.5 \\
   & $\checkmark$   & 7.7  &  4.3 &  60.9  & \textbf{35.2} & 15.7 & 26.3 \\
\bottomrule \\
\end{tabular}
\caption{Performance of Transfusion with and without limiting the amount of sampled diffusion noise to a maximum of $t=500$ when images appear before the caption.
The models are U-Net variants encoding 2$\times$2 latent pixel patches.
Metrics that change by over 1\% are bolded.}
\label{tab:noise_cap}
\end{table}

\subsection{Comparison with Image Generation Literature}
\label{sec:enhanced}

Our experiments thus far have covered controlled comparisons with Chameleon and Llama, but we have yet to compare Transfusion's image generation capabilities to those of state-of-the-art image generation models.
To that end, we train a 7B parameter model with U-Net encoding/decoding layers (2$\times$2 latent pixel patches) over the equivalent of 2T tokens, comprising of 1T text corpus tokens and 3.5B images and their captions.
While the Transfusion variant in \S\ref{sec:simple} favored simplicity and experimental control, the design choices and data mixture (\S\ref{sec:setup}) of this variant lean a bit more towards image generation.
Figure~\ref{fig:samples1} and Appendix~\ref{sec:examples_image_generation} showcase generated images from this model.

We compare the performance of our model to reported results of other similar scale image generation models, as well as some publicly available text generating models for reference.
Table~\ref{tab:enhanced-joint} shows that Transfusion achieves similar performance to high-performing image generation models such as DeepFloyd \citep{deepfloyd}, while surpassing previously published models including SDXL \citep{podell2023sdxl}.
While Transfusion does lag behind SD 3 \citep{esser2024scaling}, this model leveraged synthetic image captions through backtranslation \citep{dalle3}, which enhances its GenEval performance by 6.5\% absolute (0.433$\rightarrow$0.498) at smaller scale; for simplicity, our experimental setup only included natural data.
Finally, we note that our Transfusion model can also generate text, and performs on par with the Llama models, which were trained on the same text data distribution (\S\ref{sec:setup}).

\begin{table}[t]
\small
\centering
\begin{tabular}{@{}lcccccc@{}}
\toprule
\multirow{2}{*}{\textbf{Model}} & \multicolumn{1}{c}{\textbf{Model}} & \multicolumn{1}{c}{\textbf{Text}} & \multirow{2}{*}{\textbf{Images}} & \multicolumn{1}{c}{\textbf{Llama}} & \multicolumn{1}{c}{\textbf{COCO}} & \multicolumn{1}{c}{\textbf{Gen}} \\
& \textbf{Params} & \textbf{Tokens} & & \textbf{Acc} ($\uparrow$) & \textbf{FID} ($\downarrow$) & \textbf{Eval} ($\uparrow$) \\
\midrule
Llama 1 \citep{touvron2023llama} & 7B & 1.4T & --- & 66.1 & --- & --- \\
Llama 2 \citep{llama2} & 7B & 2.0T & --- & 66.3 & --- & --- \\
Chameleon \citep{team2024chameleon} & 7B & 6.0T & 3.5B & 67.1 & 26.74~~ & 0.39 \\
\midrule
Imagen \citep{saharia2022photorealistic} & 2.6B + 4.7B$^{*}$ & --- & 5.0B & --- & 7.27 & --- \\
Parti \citep{yu2022scaling} & 20B & --- & 4.8B & --- & $^{r}$7.23~~ & --- \\
SD 1.5 \citep{rombach2022high} & 0.9B + 0.1B$^{*}$ & --- & 4.0B & --- & --- & 0.43 \\
SD 2.1 \citep{rombach2022high} & 0.9B + 0.1B$^{*}$ & --- & 2.3B & --- & --- & 0.50 \\
DALL-E 2 \citep{dalle2diffusion} & 4.2B + 1B$^{*}$ & --- & 2.6B & --- & 10.39~~ & 0.52 \\
SDXL \citep{podell2023sdxl} & 2.6B + 0.8B$^*$ & --- & 1.6B & --- & --- & 0.55 \\
DeepFloyd \citep{deepfloyd} & 5.5B + 4.7B$^{*}$ & --- & 7.5B & --- & 6.66 & 0.61 \\
SD 3 \citep{sd3} & 8B + 4.7B$^{*}$& --- & $^{s}$2.0B~~ & --- & --- & 0.68 \\
\midrule
Transfusion (Ours) & 7.3B & 1.0T & 3.5B & 66.1 & 6.78 & 0.63 \\
\bottomrule \\
\end{tabular}
\caption{Performance of a 7B Transfusion model (U-Net encoder/decoder layers, 2$\times$2 latent pixel patches) trained on the equivalent of 2T tokens, compared to similar scale models in the literature.
Except Chameleon, all the other models are restricted to generating one modality (either text or image).
$^{*}$ Frozen text encoder parameters.
$^{r}$ Parti samples 16 images for every prompt and then reranks with an auxiliary scoring model.
$^s$ SD 3 trains with synthetic caption data, which provides boosts GenEval performance.
}
\label{tab:enhanced-joint}
\end{table}

\subsection{Image Editing}
\label{sec:editing}

Our Transfusion models, which have been pretrained on text-text, image-text, and text-image data, perform well across these modality pairings.
Can these models extend their capabilities to generate images based on other images?
To investigate, we fine-tuned our 7B model (\S\ref{sec:enhanced}) using a dataset of only 8k publicly available image editing examples, where each example consists of an input image, an edit prompt, and an output image.
This approach, inspired by LIMA \citep{zhou2024lima}, allows us to assess how well the model can generalize to image-to-image generation, a scenario not covered during pretraining.

\begin{figure}[t]
    \centering
    \subfloat[Remove the cupcake on the plate.]{\includegraphics[width=0.23\textwidth]{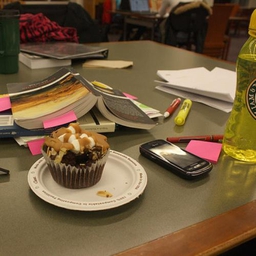}\includegraphics[width=0.23\textwidth]{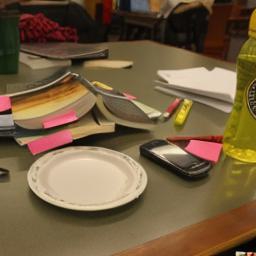}}\hfill
    \subfloat[Change the tomato on the right to a green olive.]{\includegraphics[width=0.23\textwidth]{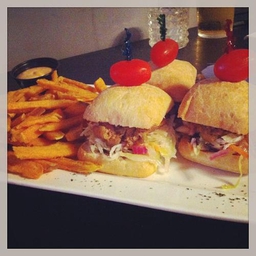}\includegraphics[width=0.23\textwidth]{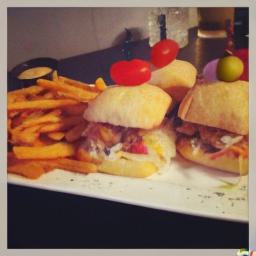}}\\[10pt]
    
    \subfloat[Write the word "Zebra" in Arial bold.]{\includegraphics[width=0.23\textwidth]{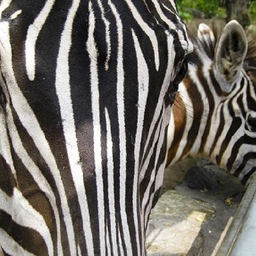}\includegraphics[width=0.23\textwidth]{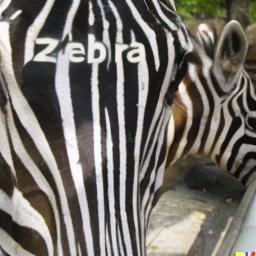}}\hfill
    \subfloat[Change this to cartoon style.]{\includegraphics[width=0.23\textwidth]{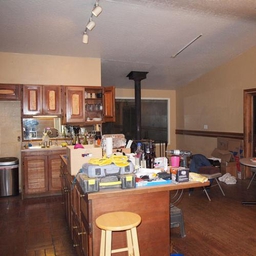}\includegraphics[width=0.23\textwidth]{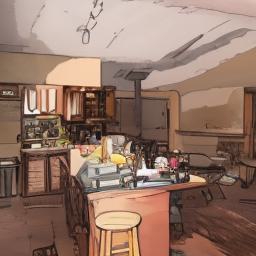}}\\[10pt]
\caption{Edited images from a fine-tuned 7B Transfusion model.}
\label{fig:edit1}
\end{figure}

Manual examination of random examples from the EmuEdit test set \citep{sheynin2024emu}, shown in Figure~\ref{fig:edit1} and Appendix~\ref{sec:editing}, reveals that our fine-tuned Transfusion model performs image edits as instructed.
Despite the limitations of this experiment, the findings suggest that Transfusion models can indeed adapt to and generalize across new modality combinations.
We leave further exploration of this promising direction to future research.

\captionsetup[subfigure]{labelformat=empty}

\section{Related Work}

Most existing multi-modal models are built on the idea of attaching two or more modality-specific architectures together, often pretraining each component separately in advance.
State-of-the-art image and video generation models, for instance, use large pretrained text encoders to represent their input prompts in latent space, which can then be used to condition diffusion models \citep{saharia2022photorealistic}.
In fact, recent work fuses representations from multiple off-the-shelf encoders to enhance performance \citep{podell2023sdxl,sd3}.
A similar pattern can be observed in the vision language model literature, where typically a pretrained language model is complemented by pretrained modality-specific encoders/decoders via projection layers to/from the pretrained text space.
Examples include Flamingo \citep{alayrac2022flamingo} and LLaVA \citep{liu2024visual} for visual understanding, GILL \citep{koh2024generating} for visual generation, and DreamLLM \citep{dongdreamllm} for both visual comprehension and generation.
In contrast, Transfusion has one unified architecture learned end-to-end to generate both text and images.

Prior work on end-to-end multi-modal models includes examples such as Fuyu \citep{fuyu-8b}, which uses image patches as inputs for visual understanding, and Chameleon \citep{team2024chameleon}, which converts each image to a sequence of discretized tokens and then trains over the combined text-image token sequences.
However, these approaches are either restricted to input-level multi-modal tasks, or lag behind state-of-the-art models (i.e. diffusion models) in continuous data generation.
Transfusion provides a simple, end-to-end solution to multi-modal learning that understands and generates high-quality multi-modal data.

An interesting area of recent acrive research is the application diffusion models and their generalizations to discrete text generation \citep{Li-2022-DiffusionLM, gat2024discrete}.
However, this approach has yet to achieve the performance and scale of standard autoregressive language models.
Future research in this direction may unlock new ways to fuse discrete and continuous modalities in a single model.

\section{Conclusion}

This work explores how to bridge the gap between the state of the art in discrete sequence modeling (next token prediction) and continuous media generation (diffusion).
We propose a simple, yet previously unexplored solution: train a single joint model on two objectives, tying each modality to its preferred objective.
Our experiments show that Transfusion scales efficiently, incurring little to no parameter sharing cost, while enabling the generation of any modality.

\begin{ack}
We would like to thank Horace He, Songlin Yang, Jiatao Gu, and Ishan Misra for helpful discussions throughout this project.
\end{ack}

\bibliography{ref}

\begin{thebibliography}{54}
\providecommand{\natexlab}[1]{#1}
\providecommand{\url}[1]{\texttt{#1}}
\expandafter\ifx\csname urlstyle\endcsname\relax
  \providecommand{\doi}[1]{doi: #1}\else
  \providecommand{\doi}{doi: \begingroup \urlstyle{rm}\Url}\fi

\bibitem[Alayrac et~al.(2022)Alayrac, Donahue, Luc, Miech, Barr, Hasson, Lenc, Mensch, Millican, Reynolds, et~al.]{alayrac2022flamingo}
Jean-Baptiste Alayrac, Jeff Donahue, Pauline Luc, Antoine Miech, Iain Barr, Yana Hasson, Karel Lenc, Arthur Mensch, Katherine Millican, Malcolm Reynolds, et~al.
\newblock Flamingo: a visual language model for few-shot learning.
\newblock \emph{Advances in neural information processing systems}, 35:\penalty0 23716--23736, 2022.

\bibitem[Bar-Tal et~al.(2024)Bar-Tal, Chefer, Tov, Herrmann, Paiss, Zada, Ephrat, Hur, Li, Michaeli, et~al.]{bar2024lumiere}
Omer Bar-Tal, Hila Chefer, Omer Tov, Charles Herrmann, Roni Paiss, Shiran Zada, Ariel Ephrat, Junhwa Hur, Yuanzhen Li, Tomer Michaeli, et~al.
\newblock Lumiere: A space-time diffusion model for video generation.
\newblock \emph{arXiv preprint arXiv:2401.12945}, 2024.

\bibitem[Bavishi et~al.(2023)Bavishi, Elsen, Hawthorne, Nye, Odena, Somani, and Ta\c{s}\i{}rlar]{fuyu-8b}
Rohan Bavishi, Erich Elsen, Curtis Hawthorne, Maxwell Nye, Augustus Odena, Arushi Somani, and Sa\u{g}nak Ta\c{s}\i{}rlar.
\newblock Introducing our multimodal models, 2023.
\newblock URL \url{https://www.adept.ai/blog/fuyu-8b}.

\bibitem[Betker et~al.(2023)Betker, Goh, Jing, Brooks, Wang, Li, Ouyang, Zhuang, Lee, Guo, Manassra, Dhariwal, Chu, Jiao, and Ramesh]{dalle3}
James Betker, Gabriel Goh, Li~Jing, Tim Brooks, Jianfeng Wang, Linjie Li, Long Ouyang, Juntang Zhuang, Joyce Lee, Yufei Guo, Wesam Manassra, Prafulla Dhariwal, Casey Chu, Yunxin Jiao, and Aditya Ramesh.
\newblock Improving image generation with better captions, 2023.
\newblock URL \url{https://api.semanticscholar.org/CorpusID:264403242}.

\bibitem[Bisk et~al.(2020)Bisk, Zellers, Gao, Choi, et~al.]{bisk2020piqa}
Yonatan Bisk, Rowan Zellers, Jianfeng Gao, Yejin Choi, et~al.
\newblock Piqa: Reasoning about physical commonsense in natural language.
\newblock In \emph{Proceedings of the AAAI conference on artificial intelligence}, pages 7432--7439, 2020.

\bibitem[{Chameleon Team}(2024)]{team2024chameleon}
{Chameleon Team}.
\newblock Chameleon: Mixed-modal early-fusion foundation models.
\newblock \emph{arXiv preprint arXiv:2405.09818}, 2024.

\bibitem[Changpinyo et~al.(2021)Changpinyo, Sharma, Ding, and Soricut]{cc12m}
Soravit Changpinyo, Piyush Sharma, Nan Ding, and Radu Soricut.
\newblock Conceptual 12m: Pushing web-scale image-text pre-training to recognize long-tail visual concepts.
\newblock \emph{CoRR}, abs/2102.08981, 2021.
\newblock URL \url{https://arxiv.org/abs/2102.08981}.

\bibitem[Chen et~al.(2020)Chen, Fan, Girshick, and He]{chen2020improved}
Xinlei Chen, Haoqi Fan, Ross Girshick, and Kaiming He.
\newblock Improved baselines with momentum contrastive learning.
\newblock \emph{arXiv preprint arXiv:2003.04297}, 2020.

\bibitem[Clark et~al.(2019)Clark, Lee, Chang, Kwiatkowski, Collins, and Toutanova]{clark2019boolq}
Christopher Clark, Kenton Lee, Ming-Wei Chang, Tom Kwiatkowski, Michael Collins, and Kristina Toutanova.
\newblock Boolq: Exploring the surprising difficulty of natural yes/no questions.
\newblock \emph{arXiv preprint arXiv:1905.10044}, 2019.

\bibitem[Clark et~al.(2018)Clark, Cowhey, Etzioni, Khot, Sabharwal, Schoenick, and Tafjord]{clark2018think}
Peter Clark, Isaac Cowhey, Oren Etzioni, Tushar Khot, Ashish Sabharwal, Carissa Schoenick, and Oyvind Tafjord.
\newblock Think you have solved question answering? try arc, the ai2 reasoning challenge.
\newblock \emph{arXiv preprint arXiv:1803.05457}, 2018.

\bibitem[Dai et~al.(2023)Dai, Hou, Ma, Tsai, Wang, Wang, Zhang, Vandenhende, Wang, Dubey, et~al.]{dai2023emu}
Xiaoliang Dai, Ji~Hou, Chih-Yao Ma, Sam Tsai, Jialiang Wang, Rui Wang, Peizhao Zhang, Simon Vandenhende, Xiaofang Wang, Abhimanyu Dubey, et~al.
\newblock Emu: Enhancing image generation models using photogenic needles in a haystack.
\newblock \emph{arXiv preprint arXiv:2309.15807}, 2023.

\bibitem[Dong et~al.(2023)Dong, Han, Peng, Qi, Ge, Yang, Zhao, Sun, Zhou, Wei, et~al.]{dong2023dreamllm}
Runpei Dong, Chunrui Han, Yuang Peng, Zekun Qi, Zheng Ge, Jinrong Yang, Liang Zhao, Jianjian Sun, Hongyu Zhou, Haoran Wei, et~al.
\newblock Dreamllm: Synergistic multimodal comprehension and creation.
\newblock \emph{arXiv preprint arXiv:2309.11499}, 2023.

\bibitem[Dong et~al.(2024)Dong, Han, Peng, Qi, Ge, Yang, Zhao, Sun, Zhou, Wei, et~al.]{dongdreamllm}
Runpei Dong, Chunrui Han, Yuang Peng, Zekun Qi, Zheng Ge, Jinrong Yang, Liang Zhao, Jianjian Sun, Hongyu Zhou, Haoran Wei, et~al.
\newblock Dreamllm: Synergistic multimodal comprehension and creation.
\newblock In \emph{The Twelfth International Conference on Learning Representations}, 2024.

\bibitem[Dubey et~al.(2024)Dubey, Jauhri, Pandey, Kadian, Al-Dahle, Letman, Mathur, Schelten, Yang, Fan, Goyal, Hartshorn, Yang, Mitra, Sravankumar, Korenev, Hinsvark, Rao, Zhang, Rodriguez, Gregerson, Spataru, Roziere, Biron, Tang, Chern, Caucheteux, Nayak, Bi, Marra, McConnell, Keller, Touret, Wu, Wong, Ferrer, Nikolaidis, Allonsius, Song, Pintz, Livshits, Esiobu, Choudhary, Mahajan, Garcia-Olano, Perino, Hupkes, Lakomkin, AlBadawy, Lobanova, Dinan, Smith, Radenovic, Zhang, Synnaeve, Lee, Anderson, Nail, Mialon, Pang, Cucurell, Nguyen, Korevaar, Xu, Touvron, Zarov, Ibarra, Kloumann, Misra, Evtimov, Copet, Lee, Geffert, Vranes, Park, Mahadeokar, Shah, van~der Linde, Billock, Hong, Lee, Fu, Chi, Huang, Liu, Wang, Yu, Bitton, Spisak, Park, Rocca, Johnstun, Saxe, Jia, Alwala, Upasani, Plawiak, Li, Heafield, Stone, El-Arini, Iyer, Malik, Chiu, Bhalla, Rantala-Yeary, van~der Maaten, Chen, Tan, Jenkins, Martin, Madaan, Malo, Blecher, Landzaat, de~Oliveira, Muzzi, Pasupuleti, Singh, Paluri, Kardas, Oldham, Rita,
  Pavlova, Kambadur, Lewis, Si, Singh, Hassan, Goyal, Torabi, Bashlykov, Bogoychev, Chatterji, Duchenne, Çelebi, Alrassy, Zhang, Li, Vasic, Weng, Bhargava, Dubal, Krishnan, Koura, Xu, He, Dong, Srinivasan, Ganapathy, Calderer, Cabral, Stojnic, Raileanu, Girdhar, Patel, Sauvestre, Polidoro, Sumbaly, Taylor, Silva, Hou, Wang, Hosseini, Chennabasappa, Singh, Bell, Kim, Edunov, Nie, Narang, Raparthy, Shen, Wan, Bhosale, Zhang, Vandenhende, Batra, Whitman, Sootla, Collot, Gururangan, Borodinsky, Herman, Fowler, Sheasha, Georgiou, Scialom, Speckbacher, Mihaylov, Xiao, Karn, Goswami, Gupta, Ramanathan, Kerkez, Gonguet, Do, Vogeti, Petrovic, Chu, Xiong, Fu, Meers, Martinet, Wang, Tan, Xie, Jia, Wang, Goldschlag, Gaur, Babaei, Wen, Song, Zhang, Li, Mao, Coudert, Yan, Chen, Papakipos, Singh, Grattafiori, Jain, Kelsey, Shajnfeld, Gangidi, Victoria, Goldstand, Menon, Sharma, Boesenberg, Vaughan, Baevski, Feinstein, Kallet, Sangani, Yunus, Lupu, Alvarado, Caples, Gu, Ho, Poulton, Ryan, Ramchandani, Franco, Saraf,
  Chowdhury, Gabriel, Bharambe, Eisenman, Yazdan, James, Maurer, Leonhardi, Huang, Loyd, Paola, Paranjape, Liu, Wu, Ni, Hancock, Wasti, Spence, Stojkovic, Gamido, Montalvo, Parker, Burton, Mejia, Wang, Kim, Zhou, Hu, Chu, Cai, Tindal, Feichtenhofer, Civin, Beaty, Kreymer, Li, Wyatt, Adkins, Xu, Testuggine, David, Parikh, Liskovich, Foss, Wang, Le, Holland, Dowling, Jamil, Montgomery, Presani, Hahn, Wood, Brinkman, Arcaute, Dunbar, Smothers, Sun, Kreuk, Tian, Ozgenel, Caggioni, Guzmán, Kanayet, Seide, Florez, Schwarz, Badeer, Swee, Halpern, Thattai, Herman, Sizov, Guangyi, Zhang, Lakshminarayanan, Shojanazeri, Zou, Wang, Zha, Habeeb, Rudolph, Suk, Aspegren, Goldman, Molybog, Tufanov, Veliche, Gat, Weissman, Geboski, Kohli, Asher, Gaya, Marcus, Tang, Chan, Zhen, Reizenstein, Teboul, Zhong, Jin, Yang, Cummings, Carvill, Shepard, McPhie, Torres, Ginsburg, Wang, Wu, U, Saxena, Prasad, Khandelwal, Zand, Matosich, Veeraraghavan, Michelena, Li, Huang, Chawla, Lakhotia, Huang, Chen, Garg, A, Silva, Bell, Zhang, Guo,
  Yu, Moshkovich, Wehrstedt, Khabsa, Avalani, Bhatt, Tsimpoukelli, Mankus, Hasson, Lennie, Reso, Groshev, Naumov, Lathi, Keneally, Seltzer, Valko, Restrepo, Patel, Vyatskov, Samvelyan, Clark, Macey, Wang, Hermoso, Metanat, Rastegari, Bansal, Santhanam, Parks, White, Bawa, Singhal, Egebo, Usunier, Laptev, Dong, Zhang, Cheng, Chernoguz, Hart, Salpekar, Kalinli, Kent, Parekh, Saab, Balaji, Rittner, Bontrager, Roux, Dollar, Zvyagina, Ratanchandani, Yuvraj, Liang, Alao, Rodriguez, Ayub, Murthy, Nayani, Mitra, Li, Hogan, Battey, Wang, Maheswari, Howes, Rinott, Bondu, Datta, Chugh, Hunt, Dhillon, Sidorov, Pan, Verma, Yamamoto, Ramaswamy, Lindsay, Lindsay, Feng, Lin, Zha, Shankar, Zhang, Zhang, Wang, Agarwal, Sajuyigbe, Chintala, Max, Chen, Kehoe, Satterfield, Govindaprasad, Gupta, Cho, Virk, Subramanian, Choudhury, Goldman, Remez, Glaser, Best, Kohler, Robinson, Li, Zhang, Matthews, Chou, Shaked, Vontimitta, Ajayi, Montanez, Mohan, Kumar, Mangla, Ionescu, Poenaru, Mihailescu, Ivanov, Li, Wang, Jiang, Bouaziz,
  Constable, Tang, Wang, Wu, Wang, Xia, Wu, Gao, Chen, Hu, Jia, Qi, Li, Zhang, Zhang, Adi, Nam, Yu, Wang, Hao, Qian, He, Rait, DeVito, Rosnbrick, Wen, Yang, and Zhao]{llama3}
Abhimanyu Dubey, Abhinav Jauhri, Abhinav Pandey, Abhishek Kadian, Ahmad Al-Dahle, Aiesha Letman, Akhil Mathur, Alan Schelten, Amy Yang, Angela Fan, Anirudh Goyal, Anthony Hartshorn, Aobo Yang, Archi Mitra, Archie Sravankumar, Artem Korenev, Arthur Hinsvark, Arun Rao, Aston Zhang, Aurelien Rodriguez, Austen Gregerson, Ava Spataru, Baptiste Roziere, Bethany Biron, Binh Tang, Bobbie Chern, Charlotte Caucheteux, Chaya Nayak, Chloe Bi, Chris Marra, Chris McConnell, Christian Keller, Christophe Touret, Chunyang Wu, Corinne Wong, Cristian~Canton Ferrer, Cyrus Nikolaidis, Damien Allonsius, Daniel Song, Danielle Pintz, Danny Livshits, David Esiobu, Dhruv Choudhary, Dhruv Mahajan, Diego Garcia-Olano, Diego Perino, Dieuwke Hupkes, Egor Lakomkin, Ehab AlBadawy, Elina Lobanova, Emily Dinan, Eric~Michael Smith, Filip Radenovic, Frank Zhang, Gabriel Synnaeve, Gabrielle Lee, Georgia~Lewis Anderson, Graeme Nail, Gregoire Mialon, Guan Pang, Guillem Cucurell, Hailey Nguyen, Hannah Korevaar, Hu~Xu, Hugo Touvron, Iliyan Zarov,
  Imanol~Arrieta Ibarra, Isabel Kloumann, Ishan Misra, Ivan Evtimov, Jade Copet, Jaewon Lee, Jan Geffert, Jana Vranes, Jason Park, Jay Mahadeokar, Jeet Shah, Jelmer van~der Linde, Jennifer Billock, Jenny Hong, Jenya Lee, Jeremy Fu, Jianfeng Chi, Jianyu Huang, Jiawen Liu, Jie Wang, Jiecao Yu, Joanna Bitton, Joe Spisak, Jongsoo Park, Joseph Rocca, Joshua Johnstun, Joshua Saxe, Junteng Jia, Kalyan~Vasuden Alwala, Kartikeya Upasani, Kate Plawiak, Ke~Li, Kenneth Heafield, Kevin Stone, Khalid El-Arini, Krithika Iyer, Kshitiz Malik, Kuenley Chiu, Kunal Bhalla, Lauren Rantala-Yeary, Laurens van~der Maaten, Lawrence Chen, Liang Tan, Liz Jenkins, Louis Martin, Lovish Madaan, Lubo Malo, Lukas Blecher, Lukas Landzaat, Luke de~Oliveira, Madeline Muzzi, Mahesh Pasupuleti, Mannat Singh, Manohar Paluri, Marcin Kardas, Mathew Oldham, Mathieu Rita, Maya Pavlova, Melanie Kambadur, Mike Lewis, Min Si, Mitesh~Kumar Singh, Mona Hassan, Naman Goyal, Narjes Torabi, Nikolay Bashlykov, Nikolay Bogoychev, Niladri Chatterji, Olivier
  Duchenne, Onur Çelebi, Patrick Alrassy, Pengchuan Zhang, Pengwei Li, Petar Vasic, Peter Weng, Prajjwal Bhargava, Pratik Dubal, Praveen Krishnan, Punit~Singh Koura, Puxin Xu, Qing He, Qingxiao Dong, Ragavan Srinivasan, Raj Ganapathy, Ramon Calderer, Ricardo~Silveira Cabral, Robert Stojnic, Roberta Raileanu, Rohit Girdhar, Rohit Patel, Romain Sauvestre, Ronnie Polidoro, Roshan Sumbaly, Ross Taylor, Ruan Silva, Rui Hou, Rui Wang, Saghar Hosseini, Sahana Chennabasappa, Sanjay Singh, Sean Bell, Seohyun~Sonia Kim, Sergey Edunov, Shaoliang Nie, Sharan Narang, Sharath Raparthy, Sheng Shen, Shengye Wan, Shruti Bhosale, Shun Zhang, Simon Vandenhende, Soumya Batra, Spencer Whitman, Sten Sootla, Stephane Collot, Suchin Gururangan, Sydney Borodinsky, Tamar Herman, Tara Fowler, Tarek Sheasha, Thomas Georgiou, Thomas Scialom, Tobias Speckbacher, Todor Mihaylov, Tong Xiao, Ujjwal Karn, Vedanuj Goswami, Vibhor Gupta, Vignesh Ramanathan, Viktor Kerkez, Vincent Gonguet, Virginie Do, Vish Vogeti, Vladan Petrovic, Weiwei Chu,
  Wenhan Xiong, Wenyin Fu, Whitney Meers, Xavier Martinet, Xiaodong Wang, Xiaoqing~Ellen Tan, Xinfeng Xie, Xuchao Jia, Xuewei Wang, Yaelle Goldschlag, Yashesh Gaur, Yasmine Babaei, Yi~Wen, Yiwen Song, Yuchen Zhang, Yue Li, Yuning Mao, Zacharie~Delpierre Coudert, Zheng Yan, Zhengxing Chen, Zoe Papakipos, Aaditya Singh, Aaron Grattafiori, Abha Jain, Adam Kelsey, Adam Shajnfeld, Adithya Gangidi, Adolfo Victoria, Ahuva Goldstand, Ajay Menon, Ajay Sharma, Alex Boesenberg, Alex Vaughan, Alexei Baevski, Allie Feinstein, Amanda Kallet, Amit Sangani, Anam Yunus, Andrei Lupu, Andres Alvarado, Andrew Caples, Andrew Gu, Andrew Ho, Andrew Poulton, Andrew Ryan, Ankit Ramchandani, Annie Franco, Aparajita Saraf, Arkabandhu Chowdhury, Ashley Gabriel, Ashwin Bharambe, Assaf Eisenman, Azadeh Yazdan, Beau James, Ben Maurer, Benjamin Leonhardi, Bernie Huang, Beth Loyd, Beto~De Paola, Bhargavi Paranjape, Bing Liu, Bo~Wu, Boyu Ni, Braden Hancock, Bram Wasti, Brandon Spence, Brani Stojkovic, Brian Gamido, Britt Montalvo, Carl
  Parker, Carly Burton, Catalina Mejia, Changhan Wang, Changkyu Kim, Chao Zhou, Chester Hu, Ching-Hsiang Chu, Chris Cai, Chris Tindal, Christoph Feichtenhofer, Damon Civin, Dana Beaty, Daniel Kreymer, Daniel Li, Danny Wyatt, David Adkins, David Xu, Davide Testuggine, Delia David, Devi Parikh, Diana Liskovich, Didem Foss, Dingkang Wang, Duc Le, Dustin Holland, Edward Dowling, Eissa Jamil, Elaine Montgomery, Eleonora Presani, Emily Hahn, Emily Wood, Erik Brinkman, Esteban Arcaute, Evan Dunbar, Evan Smothers, Fei Sun, Felix Kreuk, Feng Tian, Firat Ozgenel, Francesco Caggioni, Francisco Guzmán, Frank Kanayet, Frank Seide, Gabriela~Medina Florez, Gabriella Schwarz, Gada Badeer, Georgia Swee, Gil Halpern, Govind Thattai, Grant Herman, Grigory Sizov, Guangyi, Zhang, Guna Lakshminarayanan, Hamid Shojanazeri, Han Zou, Hannah Wang, Hanwen Zha, Haroun Habeeb, Harrison Rudolph, Helen Suk, Henry Aspegren, Hunter Goldman, Igor Molybog, Igor Tufanov, Irina-Elena Veliche, Itai Gat, Jake Weissman, James Geboski, James Kohli,
  Japhet Asher, Jean-Baptiste Gaya, Jeff Marcus, Jeff Tang, Jennifer Chan, Jenny Zhen, Jeremy Reizenstein, Jeremy Teboul, Jessica Zhong, Jian Jin, Jingyi Yang, Joe Cummings, Jon Carvill, Jon Shepard, Jonathan McPhie, Jonathan Torres, Josh Ginsburg, Junjie Wang, Kai Wu, Kam~Hou U, Karan Saxena, Karthik Prasad, Kartikay Khandelwal, Katayoun Zand, Kathy Matosich, Kaushik Veeraraghavan, Kelly Michelena, Keqian Li, Kun Huang, Kunal Chawla, Kushal Lakhotia, Kyle Huang, Lailin Chen, Lakshya Garg, Lavender A, Leandro Silva, Lee Bell, Lei Zhang, Liangpeng Guo, Licheng Yu, Liron Moshkovich, Luca Wehrstedt, Madian Khabsa, Manav Avalani, Manish Bhatt, Maria Tsimpoukelli, Martynas Mankus, Matan Hasson, Matthew Lennie, Matthias Reso, Maxim Groshev, Maxim Naumov, Maya Lathi, Meghan Keneally, Michael~L. Seltzer, Michal Valko, Michelle Restrepo, Mihir Patel, Mik Vyatskov, Mikayel Samvelyan, Mike Clark, Mike Macey, Mike Wang, Miquel~Jubert Hermoso, Mo~Metanat, Mohammad Rastegari, Munish Bansal, Nandhini Santhanam, Natascha
  Parks, Natasha White, Navyata Bawa, Nayan Singhal, Nick Egebo, Nicolas Usunier, Nikolay~Pavlovich Laptev, Ning Dong, Ning Zhang, Norman Cheng, Oleg Chernoguz, Olivia Hart, Omkar Salpekar, Ozlem Kalinli, Parkin Kent, Parth Parekh, Paul Saab, Pavan Balaji, Pedro Rittner, Philip Bontrager, Pierre Roux, Piotr Dollar, Polina Zvyagina, Prashant Ratanchandani, Pritish Yuvraj, Qian Liang, Rachad Alao, Rachel Rodriguez, Rafi Ayub, Raghotham Murthy, Raghu Nayani, Rahul Mitra, Raymond Li, Rebekkah Hogan, Robin Battey, Rocky Wang, Rohan Maheswari, Russ Howes, Ruty Rinott, Sai~Jayesh Bondu, Samyak Datta, Sara Chugh, Sara Hunt, Sargun Dhillon, Sasha Sidorov, Satadru Pan, Saurabh Verma, Seiji Yamamoto, Sharadh Ramaswamy, Shaun Lindsay, Shaun Lindsay, Sheng Feng, Shenghao Lin, Shengxin~Cindy Zha, Shiva Shankar, Shuqiang Zhang, Shuqiang Zhang, Sinong Wang, Sneha Agarwal, Soji Sajuyigbe, Soumith Chintala, Stephanie Max, Stephen Chen, Steve Kehoe, Steve Satterfield, Sudarshan Govindaprasad, Sumit Gupta, Sungmin Cho, Sunny
  Virk, Suraj Subramanian, Sy~Choudhury, Sydney Goldman, Tal Remez, Tamar Glaser, Tamara Best, Thilo Kohler, Thomas Robinson, Tianhe Li, Tianjun Zhang, Tim Matthews, Timothy Chou, Tzook Shaked, Varun Vontimitta, Victoria Ajayi, Victoria Montanez, Vijai Mohan, Vinay~Satish Kumar, Vishal Mangla, Vlad Ionescu, Vlad Poenaru, Vlad~Tiberiu Mihailescu, Vladimir Ivanov, Wei Li, Wenchen Wang, Wenwen Jiang, Wes Bouaziz, Will Constable, Xiaocheng Tang, Xiaofang Wang, Xiaojian Wu, Xiaolan Wang, Xide Xia, Xilun Wu, Xinbo Gao, Yanjun Chen, Ye~Hu, Ye~Jia, Ye~Qi, Yenda Li, Yilin Zhang, Ying Zhang, Yossi Adi, Youngjin Nam, Yu, Wang, Yuchen Hao, Yundi Qian, Yuzi He, Zach Rait, Zachary DeVito, Zef Rosnbrick, Zhaoduo Wen, Zhenyu Yang, and Zhiwei Zhao.
\newblock The llama 3 herd of models, 2024.
\newblock URL \url{https://arxiv.org/abs/2407.21783}.

\bibitem[Esser et~al.(2021)Esser, Rombach, and Ommer]{esser2021taming}
Patrick Esser, Robin Rombach, and Bjorn Ommer.
\newblock Taming transformers for high-resolution image synthesis.
\newblock In \emph{Proceedings of the IEEE/CVF conference on computer vision and pattern recognition}, pages 12873--12883, 2021.

\bibitem[Esser et~al.(2024{\natexlab{a}})Esser, Kulal, Blattmann, Entezari, M{\"u}ller, Saini, Levi, Lorenz, Sauer, Boesel, et~al.]{esser2024scaling}
Patrick Esser, Sumith Kulal, Andreas Blattmann, Rahim Entezari, Jonas M{\"u}ller, Harry Saini, Yam Levi, Dominik Lorenz, Axel Sauer, Frederic Boesel, et~al.
\newblock Scaling rectified flow transformers for high-resolution image synthesis.
\newblock In \emph{Forty-first International Conference on Machine Learning}, 2024{\natexlab{a}}.

\bibitem[Esser et~al.(2024{\natexlab{b}})Esser, Kulal, Blattmann, Entezari, Müller, Saini, Levi, Lorenz, Sauer, Boesel, Podell, Dockhorn, English, Lacey, Goodwin, Marek, and Rombach]{sd3}
Patrick Esser, Sumith Kulal, Andreas Blattmann, Rahim Entezari, Jonas Müller, Harry Saini, Yam Levi, Dominik Lorenz, Axel Sauer, Frederic Boesel, Dustin Podell, Tim Dockhorn, Zion English, Kyle Lacey, Alex Goodwin, Yannik Marek, and Robin Rombach.
\newblock Scaling rectified flow transformers for high-resolution image synthesis, 2024{\natexlab{b}}.
\newblock URL \url{https://arxiv.org/abs/2403.03206}.

\bibitem[Gat et~al.(2024)Gat, Remez, Shaul, Kreuk, Chen, Synnaeve, Adi, and Lipman]{gat2024discrete}
Itai Gat, Tal Remez, Neta Shaul, Felix Kreuk, Ricky~TQ Chen, Gabriel Synnaeve, Yossi Adi, and Yaron Lipman.
\newblock Discrete flow matching.
\newblock \emph{arXiv preprint arXiv:2407.15595}, 2024.

\bibitem[Ghosh et~al.(2023)Ghosh, Hajishirzi, and Schmidt]{ghosh2023geneval}
Dhruba Ghosh, Hannaneh Hajishirzi, and Ludwig Schmidt.
\newblock Geneval: An object-focused framework for evaluating text-to-image alignment.
\newblock \emph{Advances in Neural Information Processing Systems}, 36, 2023.

\bibitem[Heusel et~al.(2017)Heusel, Ramsauer, Unterthiner, Nessler, and Hochreiter]{fid}
Martin Heusel, Hubert Ramsauer, Thomas Unterthiner, Bernhard Nessler, and Sepp Hochreiter.
\newblock Gans trained by a two time-scale update rule converge to a local nash equilibrium.
\newblock \emph{Advances in neural information processing systems}, 30, 2017.

\bibitem[Ho and Salimans(2022)]{ho2022classifier}
Jonathan Ho and Tim Salimans.
\newblock Classifier-free diffusion guidance.
\newblock \emph{arXiv preprint arXiv:2207.12598}, 2022.

\bibitem[Ho et~al.(2020)Ho, Jain, and Abbeel]{ho2020denoising}
Jonathan Ho, Ajay Jain, and Pieter Abbeel.
\newblock Denoising diffusion probabilistic models.
\newblock \emph{Advances in neural information processing systems}, 33:\penalty0 6840--6851, 2020.

\bibitem[Kingma and Welling(2013)]{kingma2013auto}
Diederik~P Kingma and Max Welling.
\newblock Auto-encoding variational bayes.
\newblock \emph{arXiv preprint arXiv:1312.6114}, 2013.

\bibitem[Koh et~al.(2024)Koh, Fried, and Salakhutdinov]{koh2024generating}
Jing~Yu Koh, Daniel Fried, and Russ~R Salakhutdinov.
\newblock Generating images with multimodal language models.
\newblock \emph{Advances in Neural Information Processing Systems}, 36, 2024.

\bibitem[Li et~al.(2022)Li, Thickstun, Gulrajani, Liang, and Hashimoto]{Li-2022-DiffusionLM}
Xiang~Lisa Li, John Thickstun, Ishaan Gulrajani, Percy Liang, and Tatsunori Hashimoto.
\newblock Diffusion-lm improves controllable text generation.
\newblock \emph{ArXiv}, abs/2205.14217, 2022.

\bibitem[Lin et~al.(2014)Lin, Maire, Belongie, Hays, Perona, Ramanan, Doll{\'a}r, and Zitnick]{Eval_mscoco}
Tsung-Yi Lin, Michael Maire, Serge Belongie, James Hays, Pietro Perona, Deva Ramanan, Piotr Doll{\'a}r, and C~Lawrence Zitnick.
\newblock Microsoft coco: Common objects in context.
\newblock In \emph{European conference on computer vision}, pages 740--755. Springer, 2014.

\bibitem[Lipman et~al.(2022)Lipman, Chen, Ben-Hamu, Nickel, and Le]{lipman2022flow}
Yaron Lipman, Ricky~TQ Chen, Heli Ben-Hamu, Maximilian Nickel, and Matt Le.
\newblock Flow matching for generative modeling.
\newblock \emph{arXiv preprint arXiv:2210.02747}, 2022.

\bibitem[Liu et~al.(2024)Liu, Li, Wu, and Lee]{liu2024visual}
Haotian Liu, Chunyuan Li, Qingyang Wu, and Yong~Jae Lee.
\newblock Visual instruction tuning.
\newblock \emph{Advances in neural information processing systems}, 36, 2024.

\bibitem[Liu et~al.(2023)Liu, Cheng, Liu, Zhang, Li, Ren, Zou, Yang, Su, Zhu, et~al.]{liu2023llavaplus}
Shilong Liu, Hao Cheng, Haotian Liu, Hao Zhang, Feng Li, Tianhe Ren, Xueyan Zou, Jianwei Yang, Hang Su, Jun Zhu, et~al.
\newblock Llava-plus: Learning to use tools for creating multimodal agents.
\newblock \emph{arXiv preprint arXiv:2311.05437}, 2023.

\bibitem[Nichol and Dhariwal(2021)]{nichol2021improved}
Alexander~Quinn Nichol and Prafulla Dhariwal.
\newblock Improved denoising diffusion probabilistic models.
\newblock In \emph{International conference on machine learning}, pages 8162--8171. PMLR, 2021.

\bibitem[OpenAI et~al.(2024)OpenAI, Achiam, Adler, Agarwal, Ahmad, Akkaya, Aleman, Almeida, Altenschmidt, Altman, Anadkat, Avila, Babuschkin, Balaji, Balcom, Baltescu, Bao, Bavarian, Belgum, Bello, Berdine, Bernadett-Shapiro, Berner, Bogdonoff, Boiko, Boyd, Brakman, Brockman, Brooks, Brundage, Button, Cai, Campbell, Cann, Carey, Carlson, Carmichael, Chan, Chang, Chantzis, Chen, Chen, Chen, Chen, Chen, Chess, Cho, Chu, Chung, Cummings, Currier, Dai, Decareaux, Degry, Deutsch, Deville, Dhar, Dohan, Dowling, Dunning, Ecoffet, Eleti, Eloundou, Farhi, Fedus, Felix, Fishman, Forte, Fulford, Gao, Georges, Gibson, Goel, Gogineni, Goh, Gontijo-Lopes, Gordon, Grafstein, Gray, Greene, Gross, Gu, Guo, Hallacy, Han, Harris, He, Heaton, Heidecke, Hesse, Hickey, Hickey, Hoeschele, Houghton, Hsu, Hu, Hu, Huizinga, Jain, Jain, Jang, Jiang, Jiang, Jin, Jin, Jomoto, Jonn, Jun, Kaftan, Łukasz Kaiser, Kamali, Kanitscheider, Keskar, Khan, Kilpatrick, Kim, Kim, Kim, Kirchner, Kiros, Knight, Kokotajlo, Łukasz Kondraciuk, Kondrich,
  Konstantinidis, Kosic, Krueger, Kuo, Lampe, Lan, Lee, Leike, Leung, Levy, Li, Lim, Lin, Lin, Litwin, Lopez, Lowe, Lue, Makanju, Malfacini, Manning, Markov, Markovski, Martin, Mayer, Mayne, McGrew, McKinney, McLeavey, McMillan, McNeil, Medina, Mehta, Menick, Metz, Mishchenko, Mishkin, Monaco, Morikawa, Mossing, Mu, Murati, Murk, Mély, Nair, Nakano, Nayak, Neelakantan, Ngo, Noh, Ouyang, O'Keefe, Pachocki, Paino, Palermo, Pantuliano, Parascandolo, Parish, Parparita, Passos, Pavlov, Peng, Perelman, de~Avila Belbute~Peres, Petrov, de~Oliveira~Pinto, Michael, Pokorny, Pokrass, Pong, Powell, Power, Power, Proehl, Puri, Radford, Rae, Ramesh, Raymond, Real, Rimbach, Ross, Rotsted, Roussez, Ryder, Saltarelli, Sanders, Santurkar, Sastry, Schmidt, Schnurr, Schulman, Selsam, Sheppard, Sherbakov, Shieh, Shoker, Shyam, Sidor, Sigler, Simens, Sitkin, Slama, Sohl, Sokolowsky, Song, Staudacher, Such, Summers, Sutskever, Tang, Tezak, Thompson, Tillet, Tootoonchian, Tseng, Tuggle, Turley, Tworek, Uribe, Vallone, Vijayvergiya,
  Voss, Wainwright, Wang, Wang, Wang, Ward, Wei, Weinmann, Welihinda, Welinder, Weng, Weng, Wiethoff, Willner, Winter, Wolrich, Wong, Workman, Wu, Wu, Wu, Xiao, Xu, Yoo, Yu, Yuan, Zaremba, Zellers, Zhang, Zhang, Zhao, Zheng, Zhuang, Zhuk, and Zoph]{gpt4}
OpenAI, Josh Achiam, Steven Adler, Sandhini Agarwal, Lama Ahmad, Ilge Akkaya, Florencia~Leoni Aleman, Diogo Almeida, Janko Altenschmidt, Sam Altman, Shyamal Anadkat, Red Avila, Igor Babuschkin, Suchir Balaji, Valerie Balcom, Paul Baltescu, Haiming Bao, Mohammad Bavarian, Jeff Belgum, Irwan Bello, Jake Berdine, Gabriel Bernadett-Shapiro, Christopher Berner, Lenny Bogdonoff, Oleg Boiko, Madelaine Boyd, Anna-Luisa Brakman, Greg Brockman, Tim Brooks, Miles Brundage, Kevin Button, Trevor Cai, Rosie Campbell, Andrew Cann, Brittany Carey, Chelsea Carlson, Rory Carmichael, Brooke Chan, Che Chang, Fotis Chantzis, Derek Chen, Sully Chen, Ruby Chen, Jason Chen, Mark Chen, Ben Chess, Chester Cho, Casey Chu, Hyung~Won Chung, Dave Cummings, Jeremiah Currier, Yunxing Dai, Cory Decareaux, Thomas Degry, Noah Deutsch, Damien Deville, Arka Dhar, David Dohan, Steve Dowling, Sheila Dunning, Adrien Ecoffet, Atty Eleti, Tyna Eloundou, David Farhi, Liam Fedus, Niko Felix, Simón~Posada Fishman, Juston Forte, Isabella Fulford, Leo
  Gao, Elie Georges, Christian Gibson, Vik Goel, Tarun Gogineni, Gabriel Goh, Rapha Gontijo-Lopes, Jonathan Gordon, Morgan Grafstein, Scott Gray, Ryan Greene, Joshua Gross, Shixiang~Shane Gu, Yufei Guo, Chris Hallacy, Jesse Han, Jeff Harris, Yuchen He, Mike Heaton, Johannes Heidecke, Chris Hesse, Alan Hickey, Wade Hickey, Peter Hoeschele, Brandon Houghton, Kenny Hsu, Shengli Hu, Xin Hu, Joost Huizinga, Shantanu Jain, Shawn Jain, Joanne Jang, Angela Jiang, Roger Jiang, Haozhun Jin, Denny Jin, Shino Jomoto, Billie Jonn, Heewoo Jun, Tomer Kaftan, Łukasz Kaiser, Ali Kamali, Ingmar Kanitscheider, Nitish~Shirish Keskar, Tabarak Khan, Logan Kilpatrick, Jong~Wook Kim, Christina Kim, Yongjik Kim, Jan~Hendrik Kirchner, Jamie Kiros, Matt Knight, Daniel Kokotajlo, Łukasz Kondraciuk, Andrew Kondrich, Aris Konstantinidis, Kyle Kosic, Gretchen Krueger, Vishal Kuo, Michael Lampe, Ikai Lan, Teddy Lee, Jan Leike, Jade Leung, Daniel Levy, Chak~Ming Li, Rachel Lim, Molly Lin, Stephanie Lin, Mateusz Litwin, Theresa Lopez, Ryan
  Lowe, Patricia Lue, Anna Makanju, Kim Malfacini, Sam Manning, Todor Markov, Yaniv Markovski, Bianca Martin, Katie Mayer, Andrew Mayne, Bob McGrew, Scott~Mayer McKinney, Christine McLeavey, Paul McMillan, Jake McNeil, David Medina, Aalok Mehta, Jacob Menick, Luke Metz, Andrey Mishchenko, Pamela Mishkin, Vinnie Monaco, Evan Morikawa, Daniel Mossing, Tong Mu, Mira Murati, Oleg Murk, David Mély, Ashvin Nair, Reiichiro Nakano, Rajeev Nayak, Arvind Neelakantan, Richard Ngo, Hyeonwoo Noh, Long Ouyang, Cullen O'Keefe, Jakub Pachocki, Alex Paino, Joe Palermo, Ashley Pantuliano, Giambattista Parascandolo, Joel Parish, Emy Parparita, Alex Passos, Mikhail Pavlov, Andrew Peng, Adam Perelman, Filipe de~Avila Belbute~Peres, Michael Petrov, Henrique~Ponde de~Oliveira~Pinto, Michael, Pokorny, Michelle Pokrass, Vitchyr~H. Pong, Tolly Powell, Alethea Power, Boris Power, Elizabeth Proehl, Raul Puri, Alec Radford, Jack Rae, Aditya Ramesh, Cameron Raymond, Francis Real, Kendra Rimbach, Carl Ross, Bob Rotsted, Henri Roussez,
  Nick Ryder, Mario Saltarelli, Ted Sanders, Shibani Santurkar, Girish Sastry, Heather Schmidt, David Schnurr, John Schulman, Daniel Selsam, Kyla Sheppard, Toki Sherbakov, Jessica Shieh, Sarah Shoker, Pranav Shyam, Szymon Sidor, Eric Sigler, Maddie Simens, Jordan Sitkin, Katarina Slama, Ian Sohl, Benjamin Sokolowsky, Yang Song, Natalie Staudacher, Felipe~Petroski Such, Natalie Summers, Ilya Sutskever, Jie Tang, Nikolas Tezak, Madeleine~B. Thompson, Phil Tillet, Amin Tootoonchian, Elizabeth Tseng, Preston Tuggle, Nick Turley, Jerry Tworek, Juan Felipe~Cerón Uribe, Andrea Vallone, Arun Vijayvergiya, Chelsea Voss, Carroll Wainwright, Justin~Jay Wang, Alvin Wang, Ben Wang, Jonathan Ward, Jason Wei, CJ~Weinmann, Akila Welihinda, Peter Welinder, Jiayi Weng, Lilian Weng, Matt Wiethoff, Dave Willner, Clemens Winter, Samuel Wolrich, Hannah Wong, Lauren Workman, Sherwin Wu, Jeff Wu, Michael Wu, Kai Xiao, Tao Xu, Sarah Yoo, Kevin Yu, Qiming Yuan, Wojciech Zaremba, Rowan Zellers, Chong Zhang, Marvin Zhang, Shengjia
  Zhao, Tianhao Zheng, Juntang Zhuang, William Zhuk, and Barret Zoph.
\newblock Gpt-4 technical report, 2024.
\newblock URL \url{https://arxiv.org/abs/2303.08774}.

\bibitem[Podell et~al.(2023)Podell, English, Lacey, Blattmann, Dockhorn, M{\"u}ller, Penna, and Rombach]{podell2023sdxl}
Dustin Podell, Zion English, Kyle Lacey, Andreas Blattmann, Tim Dockhorn, Jonas M{\"u}ller, Joe Penna, and Robin Rombach.
\newblock Sdxl: Improving latent diffusion models for high-resolution image synthesis.
\newblock \emph{arXiv preprint arXiv:2307.01952}, 2023.

\bibitem[Radford et~al.(2021)Radford, Kim, Hallacy, Ramesh, Goh, Agarwal, Sastry, Askell, Mishkin, Clark, et~al.]{clip}
Alec Radford, Jong~Wook Kim, Chris Hallacy, Aditya Ramesh, Gabriel Goh, Sandhini Agarwal, Girish Sastry, Amanda Askell, Pamela Mishkin, Jack Clark, et~al.
\newblock Learning transferable visual models from natural language supervision.
\newblock \emph{arXiv preprint arXiv:2103.00020}, 2021.

\bibitem[Raffel et~al.(2019)Raffel, Shazeer, Roberts, Lee, Narang, Matena, Zhou, Li, and Liu]{t5}
Colin Raffel, Noam Shazeer, Adam Roberts, Katherine Lee, Sharan Narang, Michael Matena, Yanqi Zhou, Wei Li, and Peter~J. Liu.
\newblock Exploring the limits of transfer learning with a unified text-to-text transformer.
\newblock \emph{CoRR}, abs/1910.10683, 2019.
\newblock URL \url{http://arxiv.org/abs/1910.10683}.

\bibitem[Ramesh et~al.(2021)Ramesh, Pavlov, Goh, Gray, Voss, Radford, Chen, and Sutskever]{dalle}
Aditya Ramesh, Mikhail Pavlov, Gabriel Goh, Scott Gray, Chelsea Voss, Alec Radford, Mark Chen, and Ilya Sutskever.
\newblock Zero-shot text-to-image generation.
\newblock In \emph{International conference on machine learning}, pages 8821--8831. Pmlr, 2021.

\bibitem[Ramesh et~al.(2022)Ramesh, Dhariwal, Nichol, Chu, and Chen]{dalle2diffusion}
Aditya Ramesh, Prafulla Dhariwal, Alex Nichol, Casey Chu, and Mark Chen.
\newblock Hierarchical text-conditional image generation with clip latents, 2022.
\newblock URL \url{https://arxiv.org/abs/2204.06125}.

\bibitem[Rombach et~al.(2022{\natexlab{a}})Rombach, Blattmann, Lorenz, Esser, and Ommer]{ldm}
Robin Rombach, Andreas Blattmann, Dominik Lorenz, Patrick Esser, and Bj{\"o}rn Ommer.
\newblock High-resolution image synthesis with latent diffusion models.
\newblock In \emph{Proceedings of the IEEE/CVF conference on computer vision and pattern recognition}, pages 10684--10695, 2022{\natexlab{a}}.

\bibitem[Rombach et~al.(2022{\natexlab{b}})Rombach, Blattmann, Lorenz, Esser, and Ommer]{rombach2022high}
Robin Rombach, Andreas Blattmann, Dominik Lorenz, Patrick Esser, and Bj{\"o}rn Ommer.
\newblock High-resolution image synthesis with latent diffusion models.
\newblock In \emph{Proceedings of the IEEE/CVF conference on computer vision and pattern recognition}, pages 10684--10695, 2022{\natexlab{b}}.

\bibitem[Saharia et~al.(2022)Saharia, Chan, Saxena, Li, Whang, Denton, Ghasemipour, Gontijo~Lopes, Karagol~Ayan, Salimans, et~al.]{saharia2022photorealistic}
Chitwan Saharia, William Chan, Saurabh Saxena, Lala Li, Jay Whang, Emily~L Denton, Kamyar Ghasemipour, Raphael Gontijo~Lopes, Burcu Karagol~Ayan, Tim Salimans, et~al.
\newblock Photorealistic text-to-image diffusion models with deep language understanding.
\newblock \emph{Advances in neural information processing systems}, 35:\penalty0 36479--36494, 2022.

\bibitem[Sakaguchi et~al.(2021)Sakaguchi, Bras, Bhagavatula, and Choi]{sakaguchi2021winogrande}
Keisuke Sakaguchi, Ronan~Le Bras, Chandra Bhagavatula, and Yejin Choi.
\newblock Winogrande: An adversarial winograd schema challenge at scale.
\newblock \emph{Communications of the ACM}, 64\penalty0 (9):\penalty0 99--106, 2021.

\bibitem[Sap et~al.(2019)Sap, Rashkin, Chen, LeBras, and Choi]{sap2019socialiqa}
Maarten Sap, Hannah Rashkin, Derek Chen, Ronan LeBras, and Yejin Choi.
\newblock Socialiqa: Commonsense reasoning about social interactions.
\newblock \emph{arXiv preprint arXiv:1904.09728}, 2019.

\bibitem[Shazeer(2020)]{shazeer2020glu}
Noam Shazeer.
\newblock Glu variants improve transformer.
\newblock \emph{arXiv preprint arXiv:2002.05202}, 2020.

\bibitem[Sheynin et~al.(2024)Sheynin, Polyak, Singer, Kirstain, Zohar, Ashual, Parikh, and Taigman]{sheynin2024emu}
Shelly Sheynin, Adam Polyak, Uriel Singer, Yuval Kirstain, Amit Zohar, Oron Ashual, Devi Parikh, and Yaniv Taigman.
\newblock Emu edit: Precise image editing via recognition and generation tasks.
\newblock In \emph{Proceedings of the IEEE/CVF Conference on Computer Vision and Pattern Recognition}, pages 8871--8879, 2024.

\bibitem[{Stability AI}(2024)]{deepfloyd}
{Stability AI}.
\newblock If by deepfloyd lab at stabilityai, 2024.
\newblock URL \url{https://stability.ai/news/deepfloyd-if-text-to-image-model}.

\bibitem[Su et~al.(2024)Su, Ahmed, Lu, Pan, Bo, and Liu]{su2024roformer}
Jianlin Su, Murtadha Ahmed, Yu~Lu, Shengfeng Pan, Wen Bo, and Yunfeng Liu.
\newblock Roformer: Enhanced transformer with rotary position embedding.
\newblock \emph{Neurocomputing}, 568:\penalty0 127063, 2024.

\bibitem[Touvron et~al.(2023{\natexlab{a}})Touvron, Lavril, Izacard, Martinet, Lachaux, Lacroix, Rozi{\`e}re, Goyal, Hambro, Azhar, et~al.]{touvron2023llama}
Hugo Touvron, Thibaut Lavril, Gautier Izacard, Xavier Martinet, Marie-Anne Lachaux, Timoth{\'e}e Lacroix, Baptiste Rozi{\`e}re, Naman Goyal, Eric Hambro, Faisal Azhar, et~al.
\newblock Llama: Open and efficient foundation language models.
\newblock \emph{arXiv preprint arXiv:2302.13971}, 2023{\natexlab{a}}.

\bibitem[Touvron et~al.(2023{\natexlab{b}})Touvron, Martin, Stone, Albert, Almahairi, Babaei, Bashlykov, Batra, Bhargava, Bhosale, et~al.]{llama2}
Hugo Touvron, Louis Martin, Kevin Stone, Peter Albert, Amjad Almahairi, Yasmine Babaei, Nikolay Bashlykov, Soumya Batra, Prajjwal Bhargava, Shruti Bhosale, et~al.
\newblock Llama 2: Open foundation and fine-tuned chat models.
\newblock \emph{arXiv preprint arXiv:2307.09288}, 2023{\natexlab{b}}.

\bibitem[Van Den~Oord et~al.(2017)Van Den~Oord, Vinyals, et~al.]{vqvae}
Aaron Van Den~Oord, Oriol Vinyals, et~al.
\newblock Neural discrete representation learning.
\newblock \emph{Advances in neural information processing systems}, 30, 2017.

\bibitem[Vedantam et~al.(2015)Vedantam, Lawrence~Zitnick, and Parikh]{cider}
Ramakrishna Vedantam, C~Lawrence~Zitnick, and Devi Parikh.
\newblock Cider: Consensus-based image description evaluation.
\newblock In \emph{Proceedings of the IEEE conference on computer vision and pattern recognition}, pages 4566--4575, 2015.

\bibitem[Yu et~al.(2022)Yu, Xu, Koh, Luong, Baid, Wang, Vasudevan, Ku, et~al.]{yu2022scaling}
Jiahui Yu, Yuanzhong Xu, Jing~Yu Koh, Thang Luong, Gunjan Baid, Zirui Wang, Vijay Vasudevan, Alexander Ku, et~al.
\newblock Scaling autoregressive models for content-rich text-to-image generation.
\newblock \emph{arXiv preprint arXiv:2206.10789}, 2\penalty0 (3):\penalty0 5, 2022.

\bibitem[Yu et~al.(2023)Yu, Shi, Pasunuru, Muller, Golovneva, Wang, Babu, Tang, Karrer, Sheynin, et~al.]{yu2023scaling}
Lili Yu, Bowen Shi, Ramakanth Pasunuru, Benjamin Muller, Olga Golovneva, Tianlu Wang, Arun Babu, Binh Tang, Brian Karrer, Shelly Sheynin, et~al.
\newblock Scaling autoregressive multi-modal models: Pretraining and instruction tuning.
\newblock \emph{arXiv preprint arXiv:2309.02591}, 2023.

\bibitem[Zellers et~al.(2019)Zellers, Holtzman, Bisk, Farhadi, and Choi]{zellers2019hellaswag}
Rowan Zellers, Ari Holtzman, Yonatan Bisk, Ali Farhadi, and Yejin Choi.
\newblock Hellaswag: Can a machine really finish your sentence?
\newblock In \emph{Proceedings of the 57th Annual Meeting of the Association for Computational Linguistics (ACL-2019)}. Association for Computational Linguistics, 2019.

\bibitem[Zhang et~al.(2018)Zhang, Isola, Efros, Shechtman, and Wang]{lpips}
Richard Zhang, Phillip Isola, Alexei~A Efros, Eli Shechtman, and Oliver Wang.
\newblock The unreasonable effectiveness of deep features as a perceptual metric.
\newblock In \emph{Proceedings of the IEEE conference on computer vision and pattern recognition}, pages 586--595, 2018.

\bibitem[Zhou et~al.(2024)Zhou, Liu, Xu, Iyer, Sun, Mao, Ma, Efrat, Yu, Yu, et~al.]{zhou2024lima}
Chunting Zhou, Pengfei Liu, Puxin Xu, Srinivasan Iyer, Jiao Sun, Yuning Mao, Xuezhe Ma, Avia Efrat, Ping Yu, Lili Yu, et~al.
\newblock Lima: Less is more for alignment.
\newblock \emph{Advances in Neural Information Processing Systems}, 36, 2024.

\end{thebibliography}
\bibliographystyle{plainnat}

\newpage
\appendix
\section{Autoencoder Details}
\label{sec:vae_details}

The training objective for our VAE closely follows that of \cite{esser2021taming}:
\begin{equation*}
\mathcal{L}_{\text{VAE}} = \mathcal{L}_1 + \mathcal{L}_{\text{LPIPS}} + 0.5 \mathcal{L}_{\text{GAN}} + 0.2 \mathcal{L}_{\text{ID}} + 0.000001 \mathcal{L}_{\text{KL}}
\end{equation*}
where $L_1$ is L1 loss in pixel space, $L_{\text{LPIPS}}$ is perceptual loss based on LPIPS similarity \cite{lpips}, $L_{GAN}$ is a patch-based discriminator loss, $L_{\text{ID}}$ is a perceptual loss based on internal features of the Moco v2 model \cite{chen2020improved}, and $L_{\text{KL}}$ is the standard KL-regularization term to encourage encoder outputs towards a normal distribution.
We delay the beginning of GAN training (i.e. including the adversarial loss in the loss function) to 50,000 steps, in order to let the VAE achieve sufficiently good reconstruction performance.
We use a latent dimension of 8.

The training objective for the VQ-GAN matches that of the VAE, with one notable exception: we replace the $\mathcal{L}_{\text{KL}}$ loss with the standard codebook commitment loss $\mathcal{L}_{\text{codebook}}$ \citep{vqvae}, which encourages encoder outputs and codebook vectors to be close together.
We use $\beta = 0.25$, and use loss weighting $1.0$.
The final loss function for the VQ-VAE is therefore:
\begin{equation*}    
\mathcal{L}_{\text{VQ-VAE}} = \mathcal{L}_1 + \mathcal{L}_{\text{LPIPS}} + 0.5 \mathcal{L}_{\text{GAN}} + 0.2 \mathcal{L}_{\text{ID}} + \mathcal{L}_{\text{codebook}}
\end{equation*}
The vector quantization layer is applied after projecting the encoder outputs to 8-dimensional space.
Outside of the loss function change and the quantization layer, the training setup for the VAE (for Transfusion) and VQ-VAE (for Chameleon) are the same (e.g. same amount of training compute, same training data, and same encoder/decoder architecture).

\section{Examples: Image Generation}
\label{sec:examples_image_generation}

\begin{figure}[htp]
    \centering
    \subfloat[Downtown Seattle at sunrise. detailed ink wash.]{\includegraphics[width=0.23\textwidth]{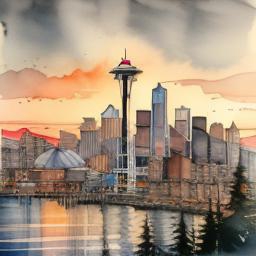}}\hfill
    \subfloat[A car made out of vegetables.]{\includegraphics[width=0.23\textwidth]{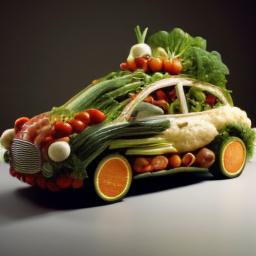}}\hfill
    \subfloat[A sign that says ``Diffusion".]{\includegraphics[width=0.23\textwidth]{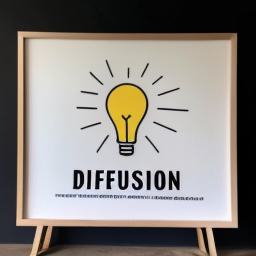}}\hfill
    \subfloat[A black basketball shoe with a lightning bolt on it.]{\includegraphics[width=0.23\textwidth]{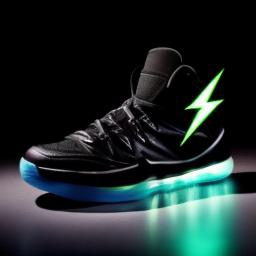}}\\[10pt]
    
    \subfloat[an espresso machine that makes coffee from human souls, high-contrast painting.]{\includegraphics[width=0.23\textwidth]{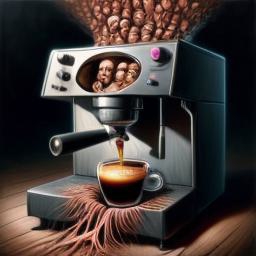}}\hfill
    \subfloat[Intricate origami of a fox and a unicorn in a snowy forest.]{\includegraphics[width=0.23\textwidth]{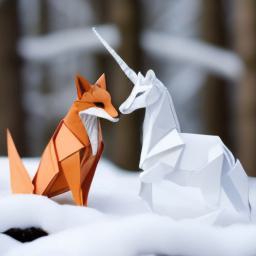}}\hfill
    \subfloat[a yellow wall with two framed sketches]{\includegraphics[width=0.23\textwidth]{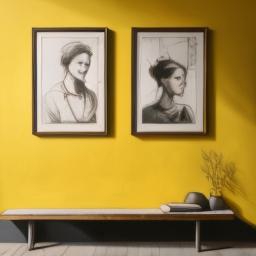}}\hfill
    \subfloat[A crab made of cheese on a plate.]{\includegraphics[width=0.23\textwidth]{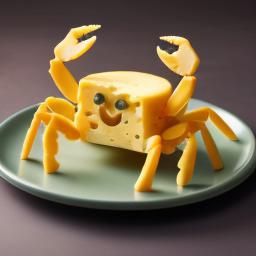}}\\[10pt]
    
    \subfloat[A single beam of light enter the room from the ceiling. The beam of light is illuminating an easel. On the easel there is a Rembrandt painting of a raccoon.]{\includegraphics[width=0.23\textwidth]{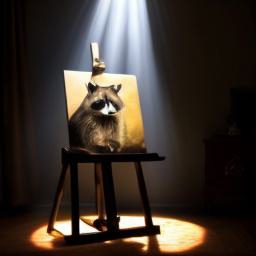}}\hfill
    \subfloat[White Cycladic houses with blue accents and vibrant magenta bougainvillea in a serene Greek island setting.]{\includegraphics[width=0.23\textwidth]{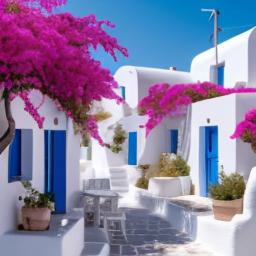}}\hfill
    \subfloat[The saying ``BE EXCELLENT TO EACH OTHER" written in a stained glass window.]{\includegraphics[width=0.23\textwidth]{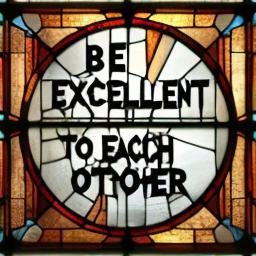}}\hfill
    \subfloat[dark high contrast render of a psychedelic tree of life illuminating dust in a mystical cave.]{\includegraphics[width=0.23\textwidth]{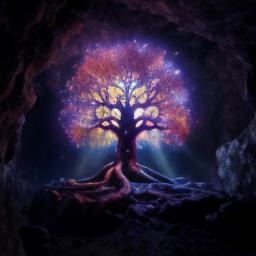}}\\[10pt]

    \subfloat[A photo of a person with the head of a cow, wearing a tuxedo and black bowtie. Beach wallpaper in the background.]{\includegraphics[width=0.23\textwidth]{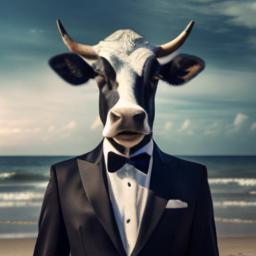}}\hfill
    \subfloat[Photo of a lychee-inspired spherical chair, with a bumpy white exterior and plush interior, set against a tropical wallpaper.]{\includegraphics[width=0.23\textwidth]{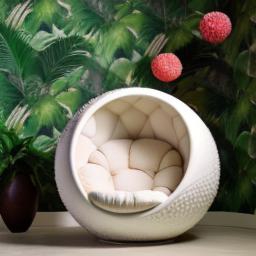}}\hfill
    \subfloat[An old rusted robot wearing pants and a jacket riding skis in a supermarket.]{\includegraphics[width=0.23\textwidth]{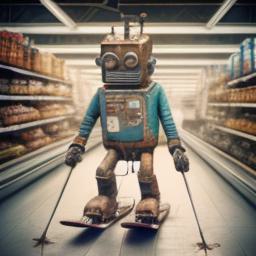}}\hfill
    \subfloat[Film still of a long-legged cute big-eye anthropomorphic cheeseburger wearing sneakers relaxing on the couch in a sparsely decorated living room.]{\includegraphics[width=0.23\textwidth]{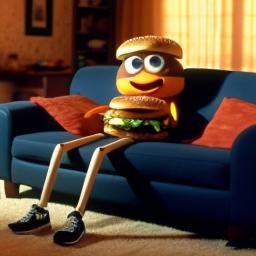}}\\[10pt]
\caption{Generated images from a 7B Transfusion trained on 2T multi-modal tokens.}
\label{fig:samples2}
\end{figure}

\begin{figure}[htp]
    \centering
    \subfloat[A woman on a bed underneath a blanket.]{\includegraphics[width=0.23\textwidth]{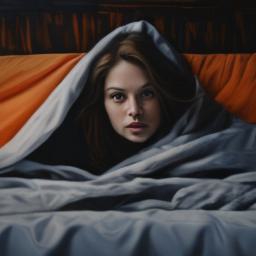}}\hfill
    \subfloat[A small blue book sitting on a large red book.]{\includegraphics[width=0.23\textwidth]{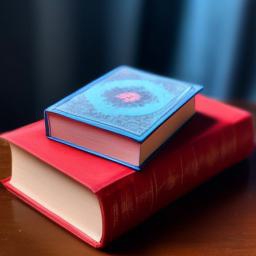}}\hfill
    \subfloat[A horse reading a book.]{\includegraphics[width=0.23\textwidth]{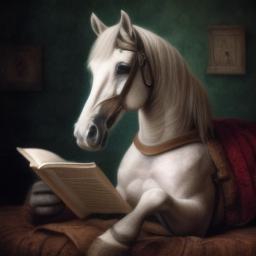}}\hfill
    \subfloat[A light bulb containing a sailboat floats through the galaxy.]{\includegraphics[width=0.23\textwidth]{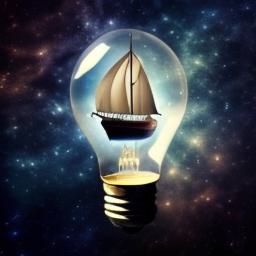}}\\[10pt]
    
    \subfloat[a monarch butterfly.]{\includegraphics[width=0.23\textwidth]{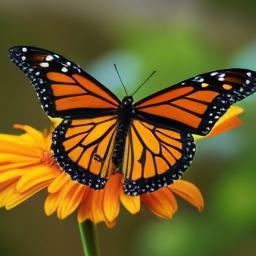}}\hfill
    \subfloat[A rowboat on a lake with a bike on it.]{\includegraphics[width=0.23\textwidth]{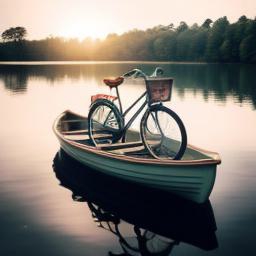}}\hfill
    \subfloat[An expressive oil painting of a chocolate chip cookie being dipped in a glass of milk, depicted as an explosion of flavors.]{\includegraphics[width=0.23\textwidth]{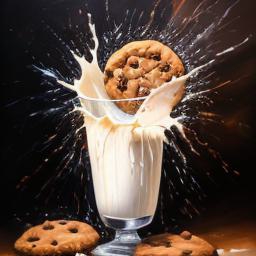}}\hfill
    \subfloat[An angry duck doing heavy weightlifting at the gym.]{\includegraphics[width=0.23\textwidth]{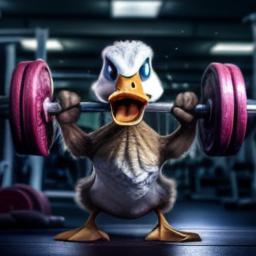}}\\[10pt]
    
    \subfloat[An emoji of a baby panda wearing a red hat, green gloves, red shirt, and green pants.]{\includegraphics[width=0.23\textwidth]{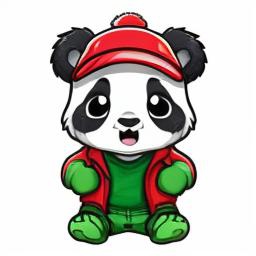}}\hfill
    \subfloat[A tranquil, anime-style koi pond in a serene Japanese garden, featuring blossoming cherry trees.]{\includegraphics[width=0.23\textwidth]{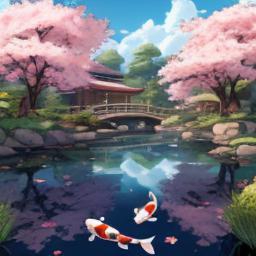}}\hfill
    \subfloat[a massive alien space ship that is shaped like a pretzel.]{\includegraphics[width=0.23\textwidth]{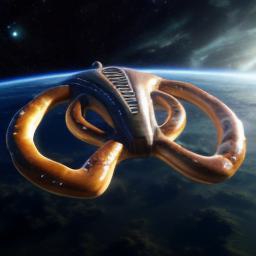}}\hfill
    \subfloat[graffiti of a funny dog on a street wall.]{\includegraphics[width=0.23\textwidth]{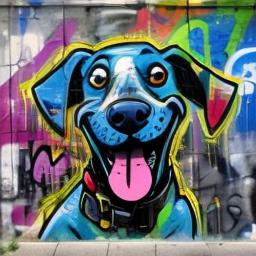}}\\[10pt]
    
    \subfloat[A spacious, serene room influenced by modern Japanese aesthetics with a view of a cityscape outside of the window.]{\includegraphics[width=0.23\textwidth]{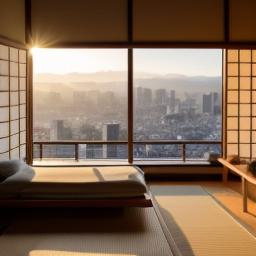}}\hfill
    \subfloat[A raccoon wearing cowboy hat and black leather jacket is behind the backyard window. Rain droplets on the window.]{\includegraphics[width=0.23\textwidth]{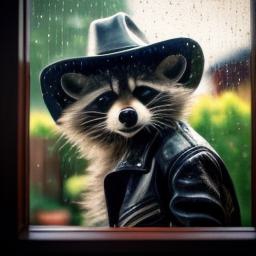}}\hfill
    \subfloat[A relaxed garlic with a blindfold reading a newspaper while floating in a pool of tomato soup.]{\includegraphics[width=0.23\textwidth]{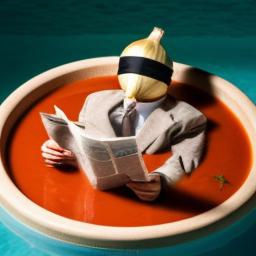}}\hfill
    \subfloat[photo of a bear wearing a suit and tophat in a river in the middle of a forest holding a sign that says ``I cant bear it".]{\includegraphics[width=0.23\textwidth]{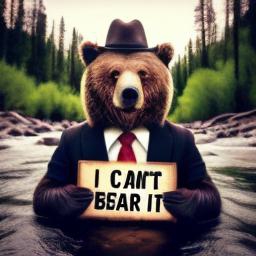}}\\[10pt]
\caption{Generated images from a 7B Transfusion trained on 2T multi-modal tokens.}
\label{fig:samples3}
\end{figure}

Figure~\ref{fig:samples2} and Figure~\ref{fig:samples3} show examples of images generated from a 7B Transfusion model trained on 2T multi-modal tokens (\S\ref{sec:enhanced}).

\section{Examples: Image Editing}
\label{sec:examples_image_editing}

\begin{figure}[htp]
    \centering
    \subfloat[Change the closest keyboard to be all black.]{\includegraphics[width=0.23\textwidth]{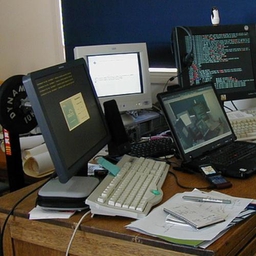}\includegraphics[width=0.23\textwidth]{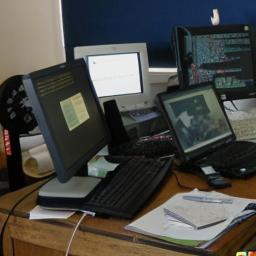}}\hfill
    \subfloat[Change the graffiti on the truck into calligraphy writing.]{\includegraphics[width=0.23\textwidth]{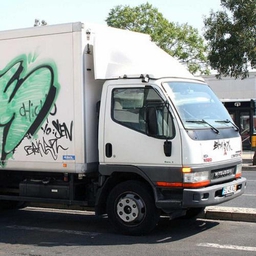}\includegraphics[width=0.23\textwidth]{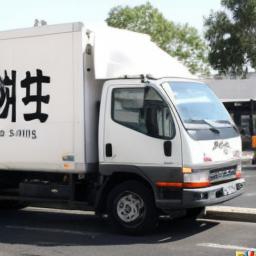}}\\[10pt]
    
    \subfloat[Can we have mountains on the background?]{\includegraphics[width=0.23\textwidth]{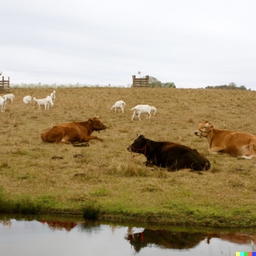}\includegraphics[width=0.23\textwidth]{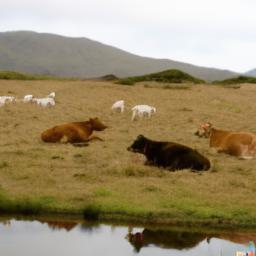}}\hfill
    \subfloat[Replace the airplane with a blackhawk helicopter.]{\includegraphics[width=0.23\textwidth]{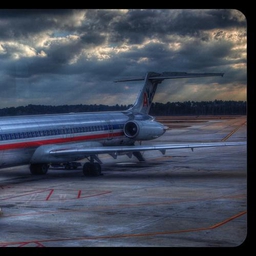}\includegraphics[width=0.23\textwidth]{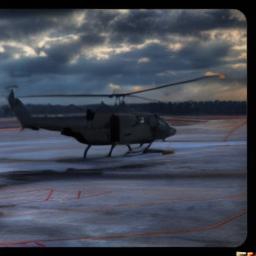}}\\[10pt]
    
    \subfloat[Add a blue rug to the floor.]{\includegraphics[width=0.23\textwidth]{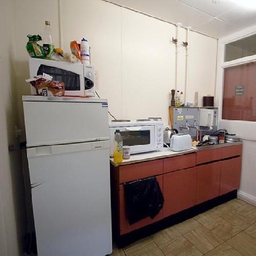}\includegraphics[width=0.23\textwidth]{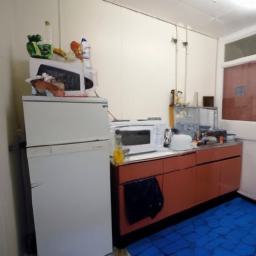}}\hfill
    \subfloat[Delete the overhead lights on top of the sink.]{\includegraphics[width=0.23\textwidth]{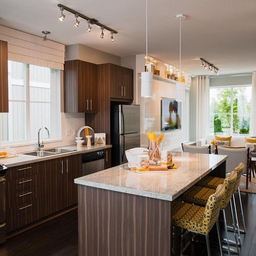}\includegraphics[width=0.23\textwidth]{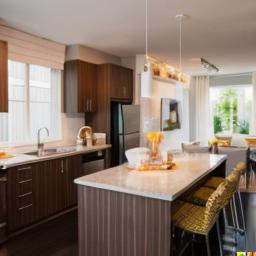}}\\[10pt]
    
    \subfloat[Change the roll of thread into a roll of wire.]{\includegraphics[width=0.23\textwidth]{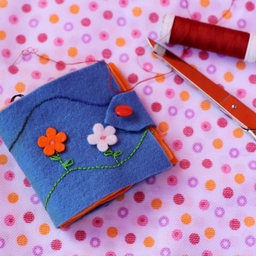}\includegraphics[width=0.23\textwidth]{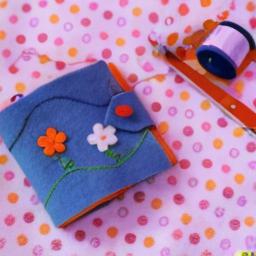}}\hfill
    \subfloat[Change the baseball bat to all brown.]{\includegraphics[width=0.23\textwidth]{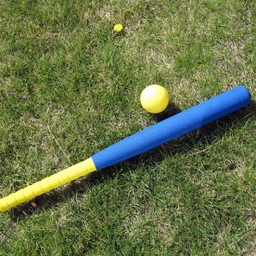}\includegraphics[width=0.23\textwidth]{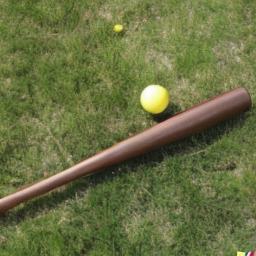}}\\[10pt]
\caption{Edited images from a fine-tuned 7B Transfusion model.}
\label{fig:edit2}
\end{figure}

Figure~\ref{fig:edit2} show random examples of image editing by a fine-tuned 7B Transfusion model.

\captionsetup[subfigure]{labelformat=empty}

\captionsetup[subfigure]{labelformat=empty}

\captionsetup[subfigure]{labelformat=empty}

\end{document}